\documentclass[acmsmall]{acmart}

\usepackage{color}
\usepackage{colortbl}
\usepackage{multirow}
\usepackage{makecell}
\usepackage{enumitem}
\usepackage{bbm}
\usepackage{amsmath}
\usepackage{threeparttable}
\usepackage{wrapfig}

\usepackage[skip=6pt]{caption}

\AtBeginDocument{%
  \providecommand\BibTeX{{%
    \normalfont B\kern-0.5em{\scshape i\kern-0.25em b}\kern-0.8em\TeX}}}

\setcopyright{acmcopyright}
\acmJournal{CSUR}
\acmPrice{15.00}\acmDOI{10.1145/3487893}

\begin{document}

\title[Statistically-Robust Clustering: A Survey]{Statistically-Robust Clustering Techniques for Mapping Spatial Hotspots: A Survey}

\author{Yiqun Xie}
\affiliation{
  \institution{University of Maryland}
  \streetaddress{Center for Geospatial Information Science, 7251 Preinkert Dr., College Park, MD 20742}
  }
\email{xie@umd.edu}

\author{Shashi Shekhar}
\affiliation{
  \institution{University of Minnesota}
  \streetaddress{Department of Computer Science, 200 Union Street SE, Minneapolis, MN 55455}
  }
\email{shekhar@umn.edu}

\author{Yan Li}
\affiliation{
	\institution{University of Minnesota}
	\streetaddress{Department of Computer Science, 200 Union Street SE, Minneapolis, MN 55455}
}
\email{lixx4266@umn.edu}

\renewcommand{\shortauthors}{Xie, Shekhar and Li}

\begin{abstract}
Mapping of spatial hotspots, i.e., regions with significantly higher rates of generating cases of certain events (e.g., disease or crime cases), is an important task in diverse societal domains, including public health, public safety, transportation, agriculture, environmental science, etc.
Clustering techniques required by these domains differ from traditional clustering methods due to the high economic and social costs of spurious results (e.g., false alarms of crime clusters). As a result, statistical rigor is needed explicitly to control the rate of spurious detections. To address this challenge, techniques for statistically-robust clustering (e.g., scan statistics) have been extensively studied by the data mining and statistics communities. In this survey we present an up-to-date and detailed review of the models and algorithms developed by this field. We first present a general taxonomy for statistically-robust clustering, covering key steps of data and statistical modeling, region enumeration and maximization, and significance testing. We further discuss different paradigms and methods within each of the key steps.
Finally, we highlight research gaps and potential future directions, which may serve as a stepping stone in generating new ideas and thoughts in this growing field and beyond. 
\end{abstract}




\maketitle

\section{Introduction}\label{sec:intro}
Identification of spatial hotspots, i.e., geographic regions with significantly higher rates of generating cases for certain types of events (e.g., disease, crime, pollution), has become an important task and standard research methodology in a diverse range of applications (e.g., public health, public safety, transportation, agriculture). 
In public health, for example, epidemiologists use spatial hotspots to find outbreaks of diseases and make timely interventions to reduce health risks to the public \cite{greene2016daily}. 
Similarly, in public safety, hotspots reveal areas with abnormally high rates of crimes (e.g., theft, assault), which can help narrow the search space for serial criminals (e.g., arsonists) or other underlying factors \cite{crime2}.
Broad applications of hotspot mapping will be detailed in Sec. \ref{sec:app}.

While spatial hotspots are naturally clusters of events, mapping of hotspots poses many unique challenges. Here we summarize these challenges in four different aspects.
\textit{First,} spurious results often have very high social and economic costs in the application domains of hotspot mapping \cite{stdmshekhar, acmkumar}. For example, if a city falsely claims a neighborhood as a crime cluster, there can be many unnecessary negative impacts such as reducing property values in that region, hurting local small businesses, increasing mental pressure on residents and potentially pushing them to move. Unfortunately, spurious clusters are very common in real-world datasets and can easily happen as a consequence of natural randomness. Thus, decision makers in important application domains require statistical rigor as a necessary component to robustly control of the rate of spurious patterns.
\textit{Second,} mapping of spatial hotspots needs explicit consideration of underlying risk or known contributing factors, which are typically not considered in traditional clustering approaches \cite{neill, satscan}. 
For example, having a hundred crime cases in a densely-populated urban area is different from having the same in a rural area of the same size. 
\textit{Third,} hotspots are geographically contiguous regions and should not be fragmented pieces in space \cite{speakman2013dynamic, takahashi2008flexibly}. As a result, spatial attributes cannot be simply mixed with other non-spatial attributes during the clustering process and need specific modeling.
\textit{Finally,} while a hotspot is often a geographically contiguous region, the observed events inside it may not necessarily form a continuous or smooth distribution of density, especially when the total number of observations is small (i.e., higher variance across locations). This undermines basic assumptions on contiguous density used in many traditional clustering approaches \cite{dbscan, campello2015hierarchical}.

Clustering has long been a core topic in data mining and machine learning, providing a vast set of techniques in various families. Partitioning-based clustering methods (e.g., k-means and Gaussian mixture models, CLARANS \cite{ng2002clarans}) split input data into $k$ groups to minimize dissimilarities within clusters.
Density-based approaches (e.g., DBSCAN \cite{dbscan}, OPTICS \cite{optics}, HDBSCAN \cite{campello2015hierarchical}) apply local or global density criteria to separate out clusters from noises. These approaches do not rely on pre-defined number of clusters and can be made flexible for different shapes and varying densities. Many techniques also utilize the hierarchical structure of clusters, e.g., by first splitting data into small units and then sequentially merging them into final clusters based on similarity and adjacency, or in the opposite direction (e.g., Chameleon \cite{karypis1999chameleon}, BIRCH \cite{zhang1996birch}, CURE \cite{guha1998cure}, HDBSCAN \cite{campello2015hierarchical}). These families of approaches are not necessarily mutually exclusive (e.g., HDBSCAN leverages both density and hierarchy).
Moving further, there have also been many other types such as grid-based (e.g., STING \cite{wang1997sting}, CLIQUE \cite{agrawal1998automatic}), graph-spectrum-based (e.g., spectral clustering \cite{von2007tutorial}), kernel-based (e.g., kernel k-means \cite{scholkopf1998nonlinear}, SVC \cite{ben2001support}), etc. Given that these techniques have been well summarized in other surveys on general clustering (e.g., \cite{xu2015comprehensive, xu2005survey}), here we skip the details and concentrate on the topic of statistically-robust clustering for hotspot mapping.

Traditional clustering techniques are not sufficient in responding to the unique challenges posed by hotspot mapping. These techniques, for example, often do not consider the high costs of spurious results and are prone to return many clusters formed by natural randomness (e.g., k-means always returns k partitions, DBSCAN and its variations have difficulty in avoiding spurious high-density regions in random point distributions \cite{sigdbscan, sigdbplus, nnscan, neill, campello2015hierarchical}). In addition, these methods normally do not consider underlying risk factors (e.g., population at risk as control), geographic-contiguity of outputs, and non-contiguous within-cluster density.

To address these challenges, in the last decades there have been a blossom of research focusing on statistically-robust clustering techniques with explicit consideration and modeling of these various needs of hotspot mapping.
As many of the methods are based on formulations from scan statistics, which was first developed by the statistics community, statistically-robust clustering is also known as scan statistics.
Since both clustering and scan statistics have been widely used in literature, we use "clustering" in the rest of the survey to better connect to intended audience from data mining.
\textit{First}, statistical significance of cluster candidates was modeled and incorporated into the detection process to provide robust control of the rate of spurious results. 
New algorithms were also developed to reduce the computational cost of significance testing, which often requires Monte-Carlo-based estimation due to the complexity of test statistics.
\textit{Second}, underlying risk factors were explicitly modeled into the test statistics 
used for candidate evaluation, allowing the scores to reflect differences in statistical processes rather than directly observed densities of events.
\textit{Third}, statistically-robust clustering techniques can also guarantee that the output clusters are contiguous in the geographic space by modeling the problem as a region-maximization problem. In this paradigm, the enumeration space of region-candidates are clearly-defined (e.g., circles, rings or linear paths formed by subsets of data) and a wide variety of efficient region-maximization algorithms 
have been developed.
This enumeration scheme also addresses the fourth challenge that within-cluster density may not be smooth or continuous due to the effect of natural randomness. By directly enumerating regions instead of forming them by local or hierarchical density criteria, regions can be evaluated regardless of their inner-density distribution.

There are several related surveys on traditional and statistically-robust clustering.
For traditional clustering, multiple taxonomies are provided in \cite{xu2015comprehensive, parimala2011survey, miller2009geographic, berkhin2006survey, xu2005survey} to summarize a broad spectrum of techniques, including partition-based, density-based, hierarchy-based, graph spectrum based, kernel based, spatial and non-spatial, etc. These surveys do not cover the models and techniques developed for statistically-robust clustering.
A few surveys also exist on statistically robust clustering from different perspectives. Kulldorff (1999) \cite{kulldorff1999spatial} provides a summary of early methods and applications of spatial scan statistics, as well as basic enumeration algorithms; an overview of earlier domain applications was also discussed later in \cite{costa2009applications}.
Neill and Moore (2006) \cite{neill2006methods} describes a few extensions of spatial scan statistic, including efficient computational strategies for rectangular-shaped clusters and an expectation-based approach to improve space-time cluster detection. Later, two sequential handbooks on scan statistics were developed mainly by the statistics community \cite{glaz2009scan,glaz2019handbook}, covering both new and existing theoretical results on distribution estimation (e.g., closed-form solutions, approximations), extreme values, sequences, etc. The handbooks also included a few chapters that are related to data mining, including summaries on Bayesian scan statistics \cite{neill2018} and irregular-shaped cluster detection \cite{duczmal2009extensions}. The topic of statistically-robust clustering (a.k.a. hotspot detection) has also been briefly discussed (e.g., as a paragraph or section) in broad spatio-temporal data mining and data science surveys \cite{acmkumar, stdmshekhar, gds}. It is worth-mentioning that hotspot is also related to another common pattern -- spatial anomaly/outlier -- in literature \cite{akoglu2015graph, chandola2009anomaly, acmkumar, stdmshekhar, gds}. One key difference between the two patterns is that anomaly or outlier is often used to describe rare observations, e.g., as defined in existing surveys such as "interesting but rare phenomena" \cite{acmkumar}, "rare occurrences" \cite{akoglu2015graph}, "rare patterns" \cite{stdmshekhar}, etc. In contrast, significant clusters or hotspots often represent a substantial/large portion or sometimes the majority of observations.

Given the richness of application contexts and the large variety of techniques developed along different dimensions of statistically-robust clustering, 
there is a need for developing a taxonomy from the computer science or data mining perspective to decompose the complex modeling and detection process into a set of key steps, highlight their roles and mutual relationships, and discuss the advances in each key step by extracting corresponding contributions from the vast (and often versatile) literature.
This can promote cross-pollination and synergistic integration of ideas across various research communities, especially considering that the developments of general clustering and statistically-robust clustering have largely run on siloed tracks. This can also help engage the broader data mining community in addressing the challenges and fostering future directions in statistically-robust clustering.

This survey aims to fulfill this need as follows. First, we develop a set of taxonomies to highlight the key components of statistically-robust clustering and its sub-areas (e.g., data models, spatial point processes, test statistics, enumeration space, computational algorithms, test paradigms). As research papers in the field often touch on multiple aspects of the overall taxonomy, we further decompose a broad set of the papers into contributions to each component of the taxonomy, which are otherwise difficult to distill and hinder cross-discipline dissemination of the research advances. This process also helps extract ideas that are applicable, and potentially valuable, to general studies in the field beyond their original scopes. Finally, based on the identified research landscape, we expose several open questions and future research directions to harness emerging sources of spatial big data and promote synergistic integration of ideas across multiple domains.

\textbf{Scope and outline:} 
This review article surveys the research progresses from a data mining perspective for computer science audience, and does not intend to provide a comprehensive survey on related statistics literature, which is another important line of research that often focuses more on statistical theories (e.g., distribution approximation) and has been well summarized in \cite{glaz2009scan,glaz2019handbook}.
The rest of the survey is outlined as follows: Sec. \ref{sec:app} introduces the application domains of hotspot mapping; Sec. \ref{sec:main} summarizes techniques for statistically-robust clustering, with Sec. \ref{sec:def} providing general key concepts, Sec. \ref{sec:taxonomy} providing an overview of building blocks, Sec. \ref{sec:data} providing a taxonomy of data models, Sec. \ref{sec:stat} providing a taxonomy on various statistical processes and test statistics, Sec. \ref{sec:region} providing a taxonomy on different region definitions as well as their enumeration and maximization algorithms, Sec. \ref{sec:test} summarizing different paradigms and methods for significance testing, and Sec. \ref{sec:eval} discussing evaluation methods and metrics. Finally, 
Sec. \ref{sec:future} and \ref{sec:conclusion} discuss challenges and opportunities for future research, and conclude the paper with highlights.

\section{Applications of Spatial Hotspot Mapping}\label{sec:app}
Statistically significant clustering for hotspot mapping has been widely used with an ever-growing variety of spatiotemporal data sources from different domains. In this section, we highlight example applications from several domains where statistically interpretable outputs and robustness against spurious results are especially important for both research investigation and policy making.

\textbf{Public health:} 
The wide adoption of hotspot mapping started from public health with the popularity of the spatial scan statistic \cite{satscan}. 
The most typical use is to identify the outbreaks of diseases, including cancer (e.g., childhood leukemia \cite{leuk}, cervical cancer \cite{chen2008geovisual}), infectious diseases (e.g., malaria \cite{disease}, dengue \cite{souza2019did, greene2016daily}, COVID-19 \cite{hohl2020daily, cordes2020spatial, ladoy2021size}), and many others \cite{ring1, greene2016daily, lord2020investigation}.
With the awareness of both diseases cases and population at risk, detected hotspots of diseases reveal abnormal underlying factors that are otherwise hidden from public health researchers and policy makers.
Many of these applications have been considered as standard practices in public health studies (e.g., National Cancer Institute \cite{nih}).
In addition to case data, other sources of related disease information, including symptoms (e.g., diarrhoea and fever \cite{multi1}) and increase in pharmacy visits \cite{bayesian1}, have been used to improve the timeliness of outbreak alarms.
With the growing variety of spatiotemporal datasets, recent applications have also explored the incorporation of proxy data from social networks (e.g., geotagged tweets about dengue fever \cite{souza2019detecting}) and trajectories.

\textbf{Public Safety:} 
Hotspot detection is popularly used in public safety applications to identify statistically alarming clusters of criminal activities, accidents, suicides, severe disaster impacts, etc.
In crime analysis, such significant clusters are used to reveal patterns of homicide and drug traffic \cite{zeoli2014homicide}, violence-related deaths \cite{minamisava2009spatial}, police deaths in the line of duty \cite{police1} and more, with respect to many underlying factors including natural disasters, weather, new construction, etc.
Geographic profiling of serial criminals, characterized as ring-shaped hotspots of serial crimes (e.g., arson), is also used to help narrow down potential residence locations of criminals \cite{ring}. 
Similarly, adoptions can be found in early warning of terrorism \cite{gao2013early}, minefield detection \cite{trang2007patterned}, 
and identification of significant concentration of accidents (e.g., falls, fire injuries, shark attacks) \cite{dey2010automated, warden2008comparison, amin2012geospatial}.

\textbf{Urban Planning and Transportation:} 
Statistically significant clustering helps urban policy makers and designers better monitor, manage and plan infrastructure changes. Typical uses include locating breakage of distribution pipes \cite{de2011detection}, mapping city areas with low accessibility to critical infrastructures or
housing difficulties \cite{perchinunno2019identification}, identifying disparities in air quality \cite{hajat2015socioeconomic}, revealing regions with aging problems \cite{xu2019investigation}, and many others \cite{helbich2012beyond, chadillon2015cannabis}. Generalized hotspots can also reflect levels of diversity (e.g., trees in urban forests, types of business services) and have been used to detect non-resilient designs or distributions of infrastructures within cities \cite{smi}.
The methods have also been extended to cover network-based applications in transportation. In transportation-safety related use cases, road segments -- including both intersections and linear paths -- have been used to identify subspaces in road networks that have significantly higher rates of traffic accidents (e.g., pedestrian fatalities) \cite{linear3, linear4}. With emerging data sources such as on-board-diagnostic data (i.e., vehicle trajectories with hundreds of engine measurements), hotspot detection is also used to find sub-paths along routes that cause divergent behaviors in emission or energy consumption \cite{li2020significant}.

\textbf{Others:} 
Forestry studies use hotspot mapping to analyze forest health at large scales \cite{riitters2005hot} and detect clusters of forest fires \cite{orozco2012cluster}, tree regeneration \cite{fei2010applying}, etc.
It is also used in both plant and animal agriculture. In plant agriculture, for example, spatial concentrations of high and low ratios of production-to-environment-cost are used to suggest alternative plans for agricultural conservation planning \cite{shackelford2015conservation}. Hotspots of crop diseases were also used to map disease spread across plants \cite{cuadros2017vector}.
On the agricultural operation safety side, the approach is used to find spatial clusters of field accidents (e.g., tractor overturns \cite{saman2012spatial}).
In animal agriculture, hotspot mapping has been widely employed to identify outbreaks of diseases in stock herds \cite{szonyi2012spatio}.
Finally, a great variety of applications exist in many other fields, including astronomy \cite{castro2018new}, medical imaging \cite{yoshida2003anatomical}, environment \cite{water1}, climate change \cite{sigchange}, fishery \cite{spindler2009spatial}, entomology \cite{bayles2017spatiotemporal}, and many more.

\section{Statistically-Robust Clustering: An Overview and Taxonomies}\label{sec:main}

In this section, we first provide key concepts and a high-level taxonomy to illustrate the building blocks of statistically-robust clustering. Then we will dive deeper into each building block and discuss the detailed taxonomy and summary of developments.

\subsection{Key concepts and a general problem statement}\label{sec:def}
In the following, we summarize the key concepts and general problem definition for statistically-robust clustering. Here we keep the definitions at a high-level given the variety of formulations when fine details are considered, which will be discussed in later sections.

\begin{definition}\textit{Case and control.}
A \textbf{case} $C_i$ is an observed incident that is related to a real-world event or phenomenon $E$ of interest (e.g., the spread of a disease). The number of cases is the number of observed incidents (e.g., the number of people with positive test results for a disease). In contrast, \textbf{control} represents the base population of candidates for the same event, where each candidate $B_j$ is subject to being a case of the event (e.g., all people being tested for a disease or at risk of being infected).
Data on case and control can be presented at either individual (e.g., points) or aggregated levels (e.g., polygons, each of which contains the number of case or control points inside).
\end{definition}

\begin{definition}\textit{Point distribution and point process.}
A \textbf{point distribution} is a spatial distribution of cases related to an event $E$. A \textbf{point process} governs the generation of random point distributions given a control distribution. In hotspot detection, it -- for example -- can determine the probability of each control point $B_j$ being a case $C_i$ of $E$ based on its location. If the spatial distribution of control is not discrete but continuous, the point process can then determine probability density across space. Here a "point" refers to a general spatial data object, which can be a real point (location), a trajectory, etc.
\end{definition}

\begin{definition}\textit{Hotspot.}
A hotspot is a sub-region of a given input space, where the rate of generating case points inside $p_{in}$ (or the probability of each individual control point being a case point) is higher compared to its outside $p_{out}$. The existence of a hotspot means the point process is not homogeneous in the input space but clustered at the hotspot region.
\end{definition}

Given case and control distributions as inputs, statistically-robust clustering aims to identify clusters formed by true hotspots rather than by pure random chance. The output clusters can either be represented using sub-regions of the input space or subsets of case points. Different from traditional clustering, output clusters are not only regions with high-intensity of case points. In contrast, the intensity is statistically adjusted by an underlying control distribution and output clusters must pass significance testing defined by a point process.

\subsection{A high-level taxonomy}\label{sec:taxonomy}

\begin{figure}
	\centering
	\includegraphics[scale=0.32]{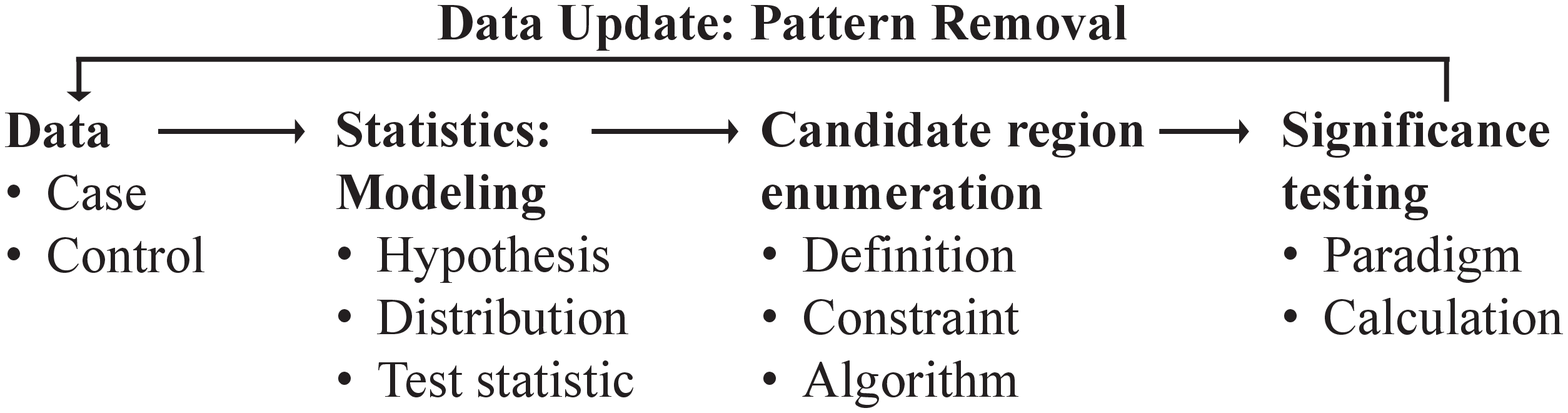}
	\caption{Taxonomy based on the process of statistically-robust clustering.}
	\centering
	\label{fig:taxonomy}
\end{figure}

Fig. \ref{fig:taxonomy} shows a high-level taxonomy of the key steps in statistically-robust clustering for hotspot mapping. Since hotspots can be considered as one type of clusters, in the rest of the paper the two names are used in an interchangeable manner.

At the beginning of the process is the input spatial data model (Fig. \ref{fig:taxonomy}). One of the key characteristics of data in hotspot mapping -- case vs. control -- is highlighted under the step. A variety of commonly used data types will be discussed and categorized in
details in Sec. \ref{sec:data}.

The second key step is the statistical modeling needed for rigorous control of spurious results. The first essential element within the step is a spatial (or spatiotemporal) point process needed to model the generation of a point distribution. This is often determined by assumptions in domain applications (e.g., Poisson point processes are common choices for monitoring disease outbreaks). Then, a hypothesis is needed to provide a clear mathematical definition of a "hotspot" (e.g., inside probability density of generating events is higher than outside). This definition is important as it spells out the criterion of a true hotspot, providing a mathematical ground to generate ground-truth synthetic data during evaluation (Sec. \ref{sec:eval}).
Finally, a test statistic is needed to compute a score of each candidate cluster, which is necessary to allow comparison and ranking among a large number of candidates.

The third key step focuses on finding "where" a hotspot is based on the statistical modeling. This is achieved by clearly defining an enumeration space (e.g., a parameter space containing all candidates) and searching over regions to maximize test statistic values. The definitions of enumeration spaces can be used to guarantee that all candidate regions are spatially contiguous and meaningful to decision makers. Additional constraints can be applied to the enumeration space to avoid undesired outputs.

The last step before a hotspot can be added to the output is significance testing. This is the final guard in this process to enforce robust control of the rate of spurious clusters. The testing can be performed using both the Frequentist and Bayesian views depending on the application need. The view should be consistent with the point process modeled in the second key step. 
Due to the typical need of Monte-Carlo-based estimation during the test, acceleration techniques are important to reduce the high cost leveraging unique characteristics of the simulation process.

Finally, in Fig. \ref{fig:taxonomy} there is an arrow connecting back to the beginning of this process, which is data update. In this step, any detected significant hotspot is removed from the data before continuing
to identify the other hotspots. This is often necessary due to the shadowing effects between hotspots in the data, which will be illustrated in Sec. \ref{sec:revisit}.

\subsection{Data types and modeling}\label{sec:data}
Mapping of hotspots covers a wide variety of data types and there are many different criteria that can be used to classify these data types into different groups. Table \ref{tab:data} shows a detailed taxonomy of various ways of classification.

\begin{table}[h]
	\centering
	\footnotesize
	\caption{Taxonomy of Data Types.}
	\label{tab:data}
	\begin{tabular}{p{2.5cm}p{3cm}p{6.5cm}}
		\hline
		\makecell[c]{\textbf{Classification}\\\textbf{criteria}} & \hfil \textbf{Categories} & \hfil \textbf{Examples and use cases}\\
		\hline
		\multirow{2}{*}{\vspace{10pt}Type of observation}
		& 
		\makecell[c]{Case}
		& 
		\vspace{-5.3pt}
		\begin{itemize}[noitemsep,nosep,topsep=0pt,,partopsep=0pt,leftmargin=*]
			\item Public health: Disease incidents (e.g., COVID), symptoms, tweets and pharmacy visits \cite{souza2019did, lord2020investigation, neill2007multivariate, hohl2020daily}
			\item Public safety: Crime cases, signals of terrorism and police deaths in the line of duty \cite{unified, ring, zeoli2014homicide, gao2013early, police1}
			\item Transportation and urban infrastructure: Vehicle accidents, pedestrian fatalities, high emissions and underground pipe breakage \cite{linear1, linear3, linear5, li2020significant, de2011detection}
			\vspace{-8pt}
		\end{itemize}\\
		\cline{2-3}
		&
		\makecell[c]{Control} &
		\vspace{-5pt}
		\begin{itemize}[noitemsep,nosep,topsep=0pt,partopsep=0pt,leftmargin=*]
			\item Public health: Population, population at risk, all tweets \cite{satscan, rect2, disease, lord2020investigation, souza2019did, souza2019detecting} 
			\item Others: Traffic volume of Vehicles or pedestrians, forest volume and crop plant population \cite{li2020significant, riitters2005hot, orozco2012cluster, fei2010applying, cuadros2017vector}\vspace{-8pt}
		\end{itemize}\\
		\hline
		\vspace{-15pt}
		\multirow{4}{*}{\vspace{14pt}\makecell[c]{Type of data models}}
		&
		\makecell[c]{Point} & 
		\vspace{-5.3pt}
		\begin{itemize}[noitemsep,nosep,topsep=0pt,partopsep=0pt,leftmargin=*]
			\item Crime cases, injuries, fires, geo-tagged tweets, sampling sites \cite{zeoli2014homicide, ring, unified, amin2012geospatial, orozco2012cluster, warden2008comparison, souza2019did, souza2019detecting, water1}
			\item Traffic accidents and vehicle tows \cite{linear2, linear3, linear4, linear5, sigdbscan}\vspace{-8pt}
		\end{itemize}
		\\
		\cline{2-3}
		&
		\makecell[c]{Polygon} & 
		\vspace{-5.3pt}
		\begin{itemize}[noitemsep,nosep,topsep=0pt,partopsep=0pt,leftmargin=*]
			\item County or city district maps with statistics of events (e.g., disease), shorelines \cite{satscan, chen2008geovisual, lord2020investigation, hohl2020daily, bayesian1, amin2012geospatial}
			\item Grid cells \cite{rect1, rect2, neill2004detecting, expectation, neill2005detection, uncertain4, sigchange, yoshida2003anatomical}\vspace{-8pt}
		\end{itemize}
		\\
		\cline{2-3}
		&
		\makecell[c]{Trajectory} & 
		\vspace{-5.3pt}
		\begin{itemize}[noitemsep,nosep,topsep=0pt,partopsep=0pt,leftmargin=*]
			\item Vehicle trajectories, smartphone or geo-tagged social media trajectories \cite{matheny2020scalable, souza2019did, souza2019detecting, li2020significant, datasource, uncertain4}\vspace{-8pt}
		\end{itemize}
		\\
		\cline{2-3}
		&
		\makecell[c]{Time-series} & 
		\vspace{-5.3pt}
		\begin{itemize}[noitemsep,nosep,topsep=0pt,partopsep=0pt,leftmargin=*]
			\item Hospital visits over the past days/weeks \cite{expectation, neill2005detection, bayesian1, bayesian2, neill2005bayesian, bayesian4}\vspace{-8pt}
		\end{itemize}
		\\
		\hline
		\multirow{2}{*}{\makecell[c]{\vspace{10pt}\hfil\space\space\space\space Type of spaces}}
		& 
		\makecell[c]{Euclidean} & 
		\vspace{-5.3pt}
		\begin{itemize}[noitemsep,nosep,topsep=0pt,partopsep=0pt,leftmargin=*]
			\item A state, city or country \cite{chen2008geovisual, lord2020investigation, hohl2020daily}
			\item A forest area \cite{orozco2012cluster, fei2010applying}\vspace{-8pt}
		\end{itemize}
		\\
		\cline{2-3}
		& 
		\makecell[c]{Spatial network or graph}& 
		\vspace{-5.3pt}
		\begin{itemize}[noitemsep,nosep,topsep=0pt,partopsep=0pt,leftmargin=*]
			\item Road or river network \cite{linear2, linear3, linear4, linear5, li2020significant}
			\item Utility network or graph representation of space \cite{de2011detection, cadena2019discovery}
			\item Social network with geo-tags \cite{souza2019detecting, souza2019did, chen2014non}\vspace{-8pt}
		\end{itemize}
		\\
		\hline
		\multirow{2}{*}{\vspace{10pt}\makecell[c]{Level of uncertainty}}
		&
		\makecell[c]{Individual} & 
		\vspace{-5.3pt}
		\begin{itemize}[noitemsep,nosep,topsep=0pt,partopsep=0pt,leftmargin=*]
			\item Locations of crimes, traffic accidents or crop disease incidents (e.g., point-type examples above)
			\item 
			Perturbed or inaccurate locations and timestamps
			\cite{uncertain1, uncertain3, uncertain4}\vspace{-8pt}
		\end{itemize}
		\\
		\cline{2-3}
		& 
		\makecell[c]{Aggregated} & 
		\vspace{-5.3pt}
		\begin{itemize}[noitemsep,nosep,topsep=0pt,partopsep=0pt,leftmargin=*]
			\item County maps with counts of cases and population, where uncertainty comes from both the aggregation and inaccuracies (e.g., \cite{uncertain6, uncertain5, cadena2018graph})\vspace{-8pt}
		\end{itemize}
		\\
		\hline
	\end{tabular}
\end{table}

\subsubsection{Types of observation}\label{sec:obs}

A key characteristic of data used in statistically-robust clustering is the need of both case and control data, which is not the case in traditional clustering approaches (i.e., no control). 
In many critical societal applications (Sec. \ref{sec:app}), 
the emergence of events are closely linked to other related distributions. In transportation, for example, a road segment with a higher traffic volume is also more likely to have a higher density of traffic accidents, but such "clusters" are not interesting to decision/policy makers as they are expected and often do not call for additional investigation and intervention \cite{linear4,li2020significant}.

To have a more holistic view of this process, we can consider two levels of an observed dataset: (1) incidents of one or more events (e.g., crime cases); and (2) underlying contributing factors to the events (i.e., control). When control is ignored, the resulting clusters are representations of an aggregation of all the underlying contributing factors -- which together make the cluster regions have higher rates of generating the observed data samples. 
Having control data as an input allows users to take out the contributions from known factors (e.g., traffic volume) -- either interesting or not -- and let resulting clusters reveal unknown factors causing higher rates of the events. More examples are provided in Table \ref{tab:data}.

\subsubsection{Types of event data models}
\begin{figure}
	\centering
	\includegraphics[scale=0.42]{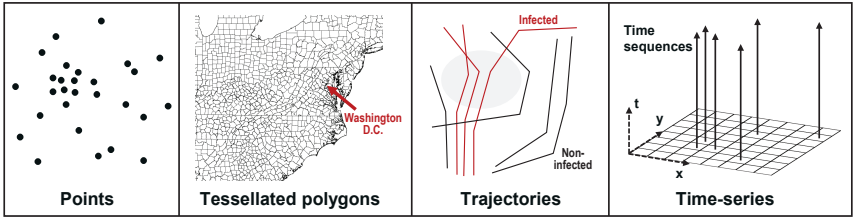}
	\caption{Example data models commonly used for hotspot mapping.}
	\centering
	\label{fig:data}
\end{figure}

Statistically-robust clustering takes a wide variety of input data models, ranging from points, polygons, grids, trajectories to time-series data. Examples of data models are shown in Fig. \ref{fig:data} as well as Table \ref{tab:data}, and illustrated as follows:

\begin{itemize}[leftmargin = *]
    \itemsep2pt 
	\item \textbf{Point:} Point is the most popular and widely used vector data model for mapping hotspots \cite{multi1,zeoli2014homicide,disease,ring1,minamisava2009spatial}. Each point has geographic coordinates, observation type (i.e., case, control), and event information (e.g., even type such as disease or crime, event subtype, event value such as air pollution indices if data is continuous). 
	\item \textbf{Polygon and grid:} Polygons are typically used when observations are given at aggregated levels \cite{value1,value2,hajat2015socioeconomic,xu2019investigation,ellip1,leuk,chen2008geovisual,greene2016daily,canccado2010penalized}. For example, population and demographic information is normally only available through census tracks, which may be shared at higher aggregation levels (e.g., census blocks or census block groups) due to privacy concerns.
	Similarly, certain types of events (e.g., cancer data) 
	may only be available at city- or county-levels.
	Such datasets are often given in the form of polygons representing boundaries of the corresponding aggregation units. 
	Some techniques take inputs in the grid format mainly for computational purposes (e.g., easier to enumerate contiguous rectangular-shaped candidate clusters) \cite{riitters2005hot,noroozi2019hotspots}. In each grid cell, a count is recorded for different types of observations and events. When original dataset provided is not in the grid format, grid-based methods will preprocess it into a grid format which may require additional assumptions for data given in polygon formats (e.g., homogeneous distribution within each polygon).
	\item \textbf{Trajectory:} As trajectory data is becoming available in greater varieties and volumes, it is also an important source of inputs for hotspot mapping. Examples of trajectory data that have been used in this field include social-media (e.g., a spatiotemporal sequence of geo-tagged tweets \cite{souza2019detecting}) or smartphone trajectories of users; location traces of vehicles and properties (e.g., taxi trajectories \cite{uncertain4, matheny2020scalable}; on-board-diagnostic data with hundreds of engine measurements \cite{li2020significant}). Trajectory data can be used in two different ways. First, it can be used to identify hotspots that have a linear-path shape in a network space (e.g., a path with abnormally high energy consumption). Second, it can be used to identify hotspots as geographic regions \cite{matheny2020scalable}, in which case events are no longer spatially static points but moving objects. This is especially valuable in epidemiology, in which a trace of locations provides much richer information than a static point (e.g., where an infected individual resides or is hospitalized) \cite{souza2019detecting, souza2019did}. One limitation of this type of data is its availability and quality (e.g., representativeness) for different types of applications.
	\item \textbf{Time-series:} In some application scenarios such as infectious disease monitoring, decision makers may be more interested in knowing about short-term dynamics reflected by the data rather than long-term phenomenons. Time-series data is very helpful for such purposes \cite{multi1}. In time-series data, a point, polygon or grid cell becomes a sequence of observations at different timestamps (e.g., numbers of visits at a hospital on a daily basis \cite{bayesian1}). This allows finding emerging hotspots, i.e., regions that were not but are becoming significant hotspots based on latest observations. 
\end{itemize}

Sometimes the input datasets are a mixture of different data models based on their availability. For example, case data may be points of crime cases while control data may be polygons (e.g., census blocks) with population information.

\subsubsection{Types of spaces}

Based on application contexts and characteristics of the phenomenon, different types of spaces may be used with input data. Two most common spaces used in hotspot mapping are Euclidean and network spaces.

Euclidean space is best suited when the goal of an application is to find geographic regions (e.g., a zone in a city) as hotspots or when the phenomenon may propagate and evolve with high-degrees of freedom (e.g., diseases that spread by air) \cite{satscan,neill2004detecting}.

Network space should be used when the goal is to find a road segment, a path or a sub-network as a hotspot \cite{de2011detection, linear2, linear3, linear5, ring2}. For example, traffic accidents mostly happen only on road networks, so a road segment or route often best represents the spatial footprint of the phenomenon. Similarly, in crime hotspot detection, if police officers are more interested in the travel distance of a serial criminal from a potential point (e.g., a residence location, which is unknown and needs to be enumerated) to a crime location \cite{ring2}, spatial networks should be used to model input data. In addition to road networks, other examples of networks include river networks, utility networks (e.g., powerlines, pipes \cite{de2011detection}), communication networks, and social-spatial networks \cite{chen2014non}.

\subsubsection{Levels of uncertainty}

Uncertainty may be introduced when data points are inaccurate either due to errors or privacy-related perturbation \cite{uncertain1, uncertain3, uncertain4}. Inputs to statistically-robust clustering may also be provided at various aggregation levels, including no-aggregation (original point samples), aggregated with nearby cases within a certain distance, aggregated at census-block or county levels, etc. Such aggregations also introduce additional uncertainty into the clustering process. The effects of these uncertainties on hotspot mapping have been examined and visualized by many \cite{uncertain1, uncertain3, uncertain4, uncertain6, uncertain5, cadena2018graph}.

\subsection{Statistical modeling}\label{sec:stat}

Statistically-robust clustering uses a variety of spatial and spatiotemporal point processes to model different types of hotspots and their corresponding hypotheses, and has developed various test statistics to evaluate hotspot candidates. In the following, we first show a taxonomy of statistical models and then discuss formulations of test statistics as well as design strategies. It is worth-noting that while many of the formulations (e.g., test statistics) are originally proposed as a part of a technique (e.g., together with maximization algorithms), they can be generally used for different problems; an analogy is the modeling of loss functions in deep learning.

\subsubsection{Spatial and Spatiotemporal Point Processes}\label{sec:process}
Point processes have been developed based on a variety of statistical models to fit different types of input data models and application needs in hotspot detection. A point process mainly concerns with point distributions over space, and the following summary includes more generalized definitions that may also consider point attributes (e.g., related to marked point processes).
\begin{itemize}[leftmargin=*]
	\item \textbf{Bernoulli and Poisson:}
	In the Bernoulli model control data (e.g., population at risk for a disease) is a collection of discrete points in some space (e.g., Euclidean, network), and 
	each point follows a Bernoulli distribution $X \sim Ber(p)$, where $p$ is the probability of being an instance of an event (e.g., disease \cite{satscan}, crime \cite{ring}, accident \cite{linear4}). 
	Statistical hypotheses are used to define hotspots (i.e., true clusters). 
	The null hypothesis $H_0$ is that all points follow the same Bernoulli distribution (same $p$) in the study area, and the alternative hypothesis $H_1$ states that there exists a hotspot, where points inside follow $Ber(p_1)$ and outside follow $Ber(p_2)$, and $p_1>p_2$.
    Instead of assuming $p$ on individual points, Poisson point processes assumes expectations at regional levels \cite{satscan}.
	\item \textbf{Multivariate:} This model takes an integrated view of multiple types of events that are related to a common parent type. Examples include multiple symptoms that relate to a single disease \cite{multi1, neill2013fast}, multiple groups in a population \cite{shen2014multivariate}, multiple actions (e.g., visiting a hospital, purchasing OTC medications) following an infection \cite{bayesian1}, and different types of crimes with similar motives. There are mainly two approaches to model the alternative hypotheses ($H_0$ still assumes a homogeneous process). The first approach considers a single $H_1$ that assumes there is a single parent event causing the increase of sub-event instances \cite{multi1}. In contrast, the second considers a set of $H_1$, and each $H_1$ relates to a different subset of sub-events \cite{bayesian1}. For example, both hospital and pharmacy visits are related to disease outbreak. However, pharmacy visits can also be caused by promotional events that are not related to an outbreak. Multiple alternative hypotheses enable such fine-grained characterization.
	\item \textbf{Multinomial:} This process also considers multiple types of events. However, these events are typically independent unlike the multivariate case \cite{multinomial, smi}. Examples of such types include races, species, unrelated diseases, etc. This modeling is used when the goal is to find sub-regions where proportions of different types differ significantly from the rest. Thus, $H_0$ states that the proportion composition is the same across the study area, whereas $H_1$ states otherwise.
	Here control data is typically not used.
	\item \textbf{Continuous:} This differs from the other processes by having continuous attributes on geo-located data points (e.g., air quality, house prices) rather than binary values or aggregated counts. To model continuous values, methods typically use normal \cite{value1,value2}, exponential \cite{huang2007spatial} or other continuous distributions \cite{jung2015nonparametric,cucala2016mann}. Using normal distribution as an example, $H_0$ states that the mean is the same across the whole study area and $H_1$ states that points in hotspot regions follow distributions with higher means.
	\item \textbf{Expectation-based:} All the processes above are mainly designed for mapping spatial hotspots but can be extended to spatiotemporal use cases.
	In contrast, the expectation-based model is designed to focus on emerging (i.e., spatiotemporal) hotspots \cite{expectation, neill2005detection}. It considers each location in a grid partition of space as a time-series of counts (e.g., number of visits to a hospital) following a Poisson process. Unlike the "inside vs. outside" view typically used in hypotheses of other processes, here it compares the most recent time period to the history of each spatial sub-region. $H_0$ states that the Poisson mean is the same for both the recent time window and past history, whereas $H_1$ says otherwise.
	\item \textbf{Bayesian:} This model also focuses on finding emerging hotspots on grid-aggregated data \cite{bayesian1,bayesian2,neill2005bayesian,bayesian4,bayesian5,bayesian6}. As an example, \cite{neill2005bayesian} uses the Bayes' rule with the conjugate Gamma-Poisson distribution to model prior and posterior, 
	where the mean of the Poisson is $p\cdot W$ where $W$ is the control population at a location and $p$ is the expected probability, and $p$ further follows a Gamma prior. $H_0$ assumes that $p$ follows the same Gamma distribution across the entire study area where $H_1$ states that there exists a sub-region where the prior is different from the rest.
\end{itemize}

\subsubsection{Test Statistics and Measures}\label{sec:measures}
In the context of statistically-robust clustering, test statistics are used to assign a score to each candidate cluster for the purpose of comparison and ranking.

Early non-probabilistic test statistics are often used to find clusters of a fixed size (e.g., $s\times s$ squares) without underlying control distribution. With these simplifying assumptions, the pure cardinality of data points in a candidate is often sufficient as a test statistic.
For more general cases with variable candidate sizes and underlying control distribution, simple extensions such as density or density ratio are used for normalization \cite{linear2, linear3, linear4}.
However, both density and density ratio are well-known to be strongly biased towards small candidates due to their monotonicity (e.g., any partitioning of a sub-region will yield a partition with a density greater than or equal to that of the sub-region). As a result, techniques based on these statistics lose detection power quickly as the size of a true hotspot increases \cite{rect2, nnscan}.

To mitigate the bias issues, most commonly-used test statistics in the frequentist framework adopted the likelihood ratio formulation (examples shown in Table \ref{tab:measure}) \cite{ring1,circle2,satscan,ellip1,rect2}.
With data generation processes being defined by a parametric distribution, the general form of the likelihood ratio for a region $S$ is $F(S) = \frac{Pr(Data|H_1(S))}{Pr(Data|H_0)}$, where $Pr(Data|H_1(S))$ and $Pr(Data|H_0)$ are the likelihoods of data being generated by the alternative hypothesis $H_1$ and null hypothesis $H_0$, respectively. A larger likelihood ratio indicates that an observation is more likely to exist under the alternative hypothesis. Different parametric distributions are used for different application scenarios (e.g., examples from Sec. \ref{sec:process}), resulting in various forms of likelihood ratios. 
For example, Poisson based likelihood ratios can be used to identify outbreaks of diseases or hotspots of crime activities \cite{ring1}; continuous versions can be used to find regions with high pollution; and expectation-based extensions are more sensitive to temporal changes \cite{expectation}.
Various studies have also explored integration of prior knowledge using regularizers in test statistics. For example, penalty terms are added to favor higher spatial resolution \cite{gangnon2004likelihood}, discourage irregularly shaped and disconnected clusters \cite{canccado2010penalized}, and promote dynamic clusters that change smoothly over time \cite{speakman2013dynamic}. 
Methods were also developed to reduce the bias of test statistics across scales (cluster sizes) and improve detection power (e.g., average likelihood ratio \cite{chan2013detection, gangnon2001weighted}, spatial mixture index \cite{smi}, nondeterministic normalization index \cite{nnscan}).
For computational efficiency, simpler formulations of test statistics are used, such as the elevated mean scan statistic \cite{qian2014connected}.
All the above-mentioned test statistics assume the underlying probability distribution obeys a parametric model. 
In scenarios where prior knowledge on underlying distributions does not exist, nonparametric model based test statistics are introduced, such as Berk Jones \cite{berk1979goodness}, nonparametric scan statistics \cite{chen2014non, neill2007nonparametric}, and Higher Criticism \cite{donoho2015special}. Without assumptions on statistical distributions, nonparametric test statistics often rely on empirical p-values for significance testing. There are also works that use a proxy to represent the underlying distributions. For example, for the continuous-value model, the Mann-Whitney statistic is used to measure the distribution of the rankings of point values instead of the actual values \cite{cucala2016mann}. Finally, these test statistics can also be applied to identify significant sub-graphs (e.g., based on Steiner-connectivity) in network or graph data as summarized in \cite{cadena2018graph}.

Many Bayesian approaches have also been developed to complement the frequentist view for significant clustering \cite{neill2018}. 
Advantages of Bayesian measures include the capacity of incorporating informative prior information, when available; the flexibility of considering multiple hypotheses; and direct calculation of posterior probabilities . 
In general, Bayesian based methods do not rely on test statistics; rather, they directly calculate posterior probabilities using the Bayes' rule (e.g., univariate \cite{neill2005bayesian, neill2006bayesian}, multivariate  \cite{neill2007multivariate, bayesian1}, PANDA-CDCA Bayesian network \cite{jiang2010bayesian, cooper2007bayesian}). With that being said, the Bayesian posterior probability can also be considered as a test statistic, and empirical thresholds can be applied as an approximation of significance levels in the frequentist setup.
There are also frequentist methods that use the Bayesian modeling to design test statistics, such as the Bayes factor \cite{zhang2008bayesian} and anomalous group detection \cite{das2009detecting}.

\begin{table}[h]
	\centering
	\footnotesize
	\caption{Examples of test statistics using point-process based likelihood ratios.}
	\label{tab:measure}
	\begin{tabular}{p{2.2cm}p{6.5cm}p{4cm}}
		\hline
		\makecell[c]{\textbf{Process types}} & \makecell[c]{\textbf{Example test statistics}} & \makecell[c]{\textbf{Interpretation}}\\
		\hline
		\makecell[c]{Poisson model}
		& 
		\vspace{-5pt}
		\hfil\begin{minipage}{6.5cm}
			\[
			\log \bigg(\frac{n}{N\cdot\frac{a}{A}} \bigg)^{n}
		\cdot
		\bigg(\frac{N-n}{N\cdot\frac{A-a}{A}} \bigg)^{N-n}
		\cdot \mathbbm{1}(n>N\cdot\frac{a}{A})
		\]
		\end{minipage}
		where $n$ and $a$ are counts of case and control in a candidate cluster, respectively; $N$ and $A$ are the total counts of case and control of the study area; and the indicator function $\mathbbm{1}(\cdot)$ guarantees the candidate is dense rather than sparse.
		&
		A log likelihood ratio (LR) using the Poisson point process. The denominators represent the expected number of cases inside and outside the candidate under $H_0$ (i.e., no hotspot) (e.g., \cite{satscan, rect2}).
		\\
		\cline{2-3}
		\makecell[c]{Continuous\\ value model}
		& 
		\vspace{-10pt}
		\hfil\begin{minipage}{6.5cm}
			\[
			N\log \sigma + \sum_{i=1}^{n}\frac{(x_i-\mu)^2}{2\sigma^2}
			-N/2-N\log \sigma'
			\]
		\end{minipage}
		where $n$ and $N$ are the number of samples of a candidate cluster and the study area, respectively; $x_i$ is the value of a sample in the candidate; $\mu$ and $\sigma$ are the mean and standard deviation of all $N$ samples; $\sigma'$ is a standard deviation estimated assuming samples inside and outside a cluster follow different means (i.e., $H_1$).
		&
		\vspace{-11.6pt}
		A log LR using the continuous normal model, where $H_0$ states all samples follow the same mean-value and $H_1$ states that there exists a candidate region following a process with a higher mean (e.g., \cite{value1, value2}).
		\\
		\cline{2-3}
		\makecell[c]{Multinomial\\ model}
		& 
		\vspace{-10pt}
		\hfil\begin{minipage}{6.5cm}
			\[
			\log \frac{\prod_{i=1}^{k} p_i^{n_i} q_i ^{N_i - n_i}}{
				\prod_{i=1}^{k} (p'_i)^{n_i} \cdot (q'_i) ^{N_i - n_i}
				}
			\]
		\end{minipage}
		where $k$ is the number of types (e.g., disease types), $n_i$ and $N_i$ are number of type-$i$ samples of a candidate cluster and the study area, respectively; $p_i = \frac{n_i}{n}$ and $q_i = \frac{N_i-n_i}{N-n}$ are the probability of a point being of type-$i$ inside and outside the candidate; and $p'_i=q'_i=N_i/N$ is the probability of a point being of type-$i$ in the whole data.
		&
		\vspace{-11.6pt}
		A log LR using the multinomial model, where $H_0$ states $p_i=q_i, \forall i = 1,...,k$ across the whole space whereas $H_1$ states some regions have different complexions (e.g., \cite{multinomial, smi}).
		\\
		\cline{2-3}
		\makecell[c]{Expectation\\based model}
		& 
		\vspace{-10pt}
		\hfil\begin{minipage}{6.5cm}
			\[
			\log \max \bigg(\big(\frac{n}{b}\big)^n \cdot e^{b-n}, \,1\bigg)
			\]
		\end{minipage}
		where $n$ is the number of cases observed in a candidate cluster at current timestamp $t$, $b$ is the baseline learned from historical data, and the $\max(\cdot,\cdot)$ function is used to filter out scenarios where $n<b$.
		&
		\vspace{-11.6pt}
		Compared to the Poisson model shown above, this log LR compares the concentration of samples in a candidate cluster at $t$ to its historical expectation $b$ from $t'<t$ rather than the concentration outside the candidate (e.g., \cite{expectation, neill2005detection}).
		\\
		\hline
	\end{tabular}
\end{table}

\subsubsection{Design strategies of test statistics and measures}\label{sec:design}
A well-designed test statistic should be unbiased and ideally have the optimal asymptotic detection power under a significance level.
Many theoretical insights and results have been developed to reveal major issues to consider during the design and recommend strategies to address them both statistically and computationally.
For example, most of existing test statistics used in research and applications are based on ratios between maximum likelihoods of hypotheses $H_1$ and $H_0$. Many of these standard formulations, including Kulldorff's likelihood ratio, have also been shown to be scale-sensitive. Specifically, these test statistics can be strongly biased towards small candidates \cite{rufibach2010block, walther2010optimal, chan2013detection, nnscan, unified} (albeit much better than density \cite{nnscan, rect1}). 

In significance testing, statistically-robust clustering techniques commonly estimate the distribution of test statistics using the best candidate (i.e., the maximum score) from the $H_0$-point-distribution in each Monte Carlo trial, which is needed to avoid multiple testing (Sec. \ref{sec:revisit}).
Thus, an intuitive interpretation behind the bias is that the distribution of the maximum score -- as a random variable -- is scale-dependent \cite{rufibach2010block, walther2010optimal, dumbgen2001multiscale, rivera2013optimal}. In other words, the mean of maximum score of smaller candidates tends to strongly dominate that of larger candidates.
As a remedy, a penalty term was constructed to correct the bias and restore the optimal asymptotic detection power for hotspots with different sizes \cite{rufibach2010block, walther2010optimal, dumbgen2001multiscale, proksch2018multiscale}.
A further improvement uses square-roots of log likelihood ratios to construct the penalty term, requiring less case-specific design \cite{rivera2013optimal}; the extension also provides additional computational efficiency (near linear-time). Another approach is to correct the bias during significance testing, where a block criterion is developed to adjust critical values (i.e., thresholds on test statistic values to remove insignificant results) on candidates belonging to different sizes or size-groups \cite{rufibach2010block, walther2010optimal}, guaranteeing optimal asymptotic power of detection.
Average likelihood ratio approaches have also been developed to reduce the bias, where each score is a weighted average of those of neighboring candidates \cite{chan2013detection, gangnon2001weighted}. According to theoretical results in \cite{chan2013detection}, the original likelihood ratio in the spatial scan statistic has optimal detection power only for hotspots with smallest footprints whereas the average likelihood ratio has optimal power at all scales. The analyses were carried out for the discrete one-dimensional scenario. Another view of the average likelihood ratio approach is that it uses marginal likelihoods instead of maximum likelihoods \cite{neill2018}, which may better utilize secondary cluster information as in the Bayesian test statistics \cite{neill2005bayesian, neill2006bayesian, neill2007multivariate}.
In addition to the pure-scale interpretation of the bias, a spatial interpretation has been developed \cite{nnscan, unified}. The analysis shows that the likelihoods or probabilities used to score a candidate in many approaches are based on a space bi-partitioning where one partition represents the candidate $C$ and the other for the outside. Likelihoods calculated using the bi-partition implicitly assume that the best candidate $C_{ran}$ in a random dataset still locates in $C$'s partition, which is rarely the case. This leads to a big underestimation especially for the likelihood of $H_0$ since the location of the best candidate should be nondeterministic in a purely random point process. To correct the bias, spatial-nondeterminism-aware test statistics were proposed \cite{nnscan, smi}, which can be computed using dual-level Monte Carlo simulation \cite{smi}. The use of another simulation introduces additional computation overhead but can be generally applied for different point processes (e.g., Bernoulli, Poisson, multinomial), test statistics (e.g., density, likelihood ratio, spatial mixture index), candidate shapes (or enumeration spaces), maximization algorithms, etc.

\subsection{Region enumeration and maximization}\label{sec:region}
Region enumeration is the core computational step in statistically robust clustering, which examines a mathematically well-defined candidate space of clusters to identify the best candidate that maximizes the test statistic for further significance testing. While this routine only returns one cluster due to maximization, techniques in the literature often repetitively apply it as a sub-routine to find arbitrary number of clusters, i.e., by removing the best cluster (if significant) from data after each round and then starting a new round for the next (e.g., \cite{satscan, rect1, smi}). If the best cluster at the current step is not significant, the algorithm terminates and returns all significant clusters. This enumeration-and-maximization (or one-cluster-at-a-time) scheme is widely used for statistically-robust clustering because of the fact that it reduces the mutual statistical influence (a.k.a. shadow effects) \cite{neill2018, li2011spatial} between multiple clusters in the same dataset (Sec. \ref{sec:revisit}).

In the following, we will first summarize definitions of enumeration spaces and constraints that are popularly used in the literature, followed by maximization algorithms as well as acceleration techniques of various types. A summary taxonomy is provided in Table \ref{tab:enum}.

\subsubsection{Enumeration Spaces}\label{sec:enumspace}
Definitions of enumeration spaces often take considerations of both application needs and computational feasibility/efficiency. These definitions can be well classified using cluster shapes (i.e., predefined, flexible), with additional consideration of data types and data spaces (e.g., Euclidean, network).

\textbf{Predefined geometric shapes:} Majority of statistically robust clustering techniques were developed for a specific type of geometric shapes (e.g., circle, ellipse, rectangle, ring, linear path) that can represent real-world phenomenons in target application domains. The corresponding enumeration space then covers all valid sub-regions of a desired shape. Circle, for example, is the most widely used shape in epidemiology and public health research \cite{satscan, diseaseold, leuk, nih, greene2016daily, lord2020investigation, circle2, matheny2020scalable} due to natural diffusion mechanisms, which tends to make infectious disease spread to an approximate circular shape; and due to computational convenience. Circles were extended to ellipses to better model effects of spatial anisotropy and provide more flexibility beyond circles \cite{ellip1}.
Rectangles were also used as proxies for circles/ellipses because of the computational convenience they offer on gridded inputs \cite{rect1, rect2}. Ring shapes were developed for applications in public safety and criminology research. According to the routine activity theory, serial criminals (e.g., arsonists) tend to commit crimes neither too close to nor too far away from their home to lower both the risk of being recognized by neighbors and the cost of travel \cite{ring}. Thus, ring-shaped zones \cite{ring, ring1, ring2} were used to depict the footprints of serial criminals. While these shapes are more natural to point data (Sec. \ref{sec:data}), they have also been used with polygon data (e.g., county polygons with total number of disease cases and population). In such cases, polygon data are first preprocessed to center points to perform region enumeration, and then in the output phase, the polygons containing points from each significant cluster are merged to generate the final results.

\begin{figure}
	\centering
	\includegraphics[scale=0.4]{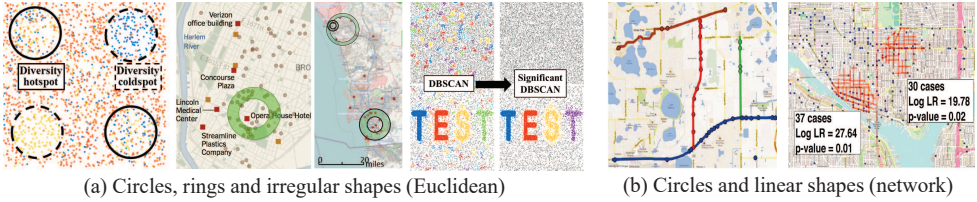}
	\caption{Example shapes used to define enumeration spaces.
	(a) Circular-shaped hotspots and coldspots of diversity \cite{smi}; ring-shaped hotspots for finding serial crimes and the source of a disease \cite{ring, ring1}; and arbitrarily-shaped hotspots by Significant DBSCAN, which also shows its robustness against spurious clusters (top: random data without clusters; bottom: clustered data) \cite{sigdbscan, sigdbplus}. (b) Linear hotspots of traffic accidents (pedestrian fatalities) \cite{linear3,linear4} and iso-distance hotspots of crimes on road networks \cite{circle2}.
}
	\centering
	\label{fig:shape}
\end{figure}

In transportation or city infrastructure related applications, vast majority of techniques use network space because data (e.g., locations of traffic accidents or pedestrian fatalities) or phenomenons are based on road, water or utility networks. Correspondingly, linear paths or intersections along network edges are used to model shapes of candidate clusters \cite{linear1, linear2, linear3, linear4, de2011detection, linear5}.
In addition, network-space has also been used to identify "areal" clusters (e.g., a circle becomes a sub-network where the network distances from all locations to a center are less than or equal to $r$), especially in urban areas where network-distance is a better measure of spatial connectivity than Euclidean distance \cite{ring2, tang2017detecting}. For example, two locations separated by a river can be very close in Euclidean space but far apart by network distance.
Fig. \ref{fig:shape} shows several common examples of shapes used to define the enumeration space.

From a computational perspective, further
simplifying constraints (e.g., two-point circles, shortest paths) 
are often needed to guarantee tractable search spaces; otherwise, a simple circular shape in a continuous space can lead to an infinite number of candidates.
This will be detailed in Sec. \ref{sec:constraint}.

\textbf{Flexible shapes:} 
A variety of techniques have been developed to detect arbitrarily shaped spatial hotspots to better capture the complex boundaries of real-world patterns caused by underlying physical and social contexts. Due to the non-parametric nature of arbitrary shapes, candidate enumeration spaces used for this paradigm are typically not well-defined. Instead, the vast majority of techniques define specific local search heuristics to iteratively go over a set of arbitrarily shaped candidates until certain criteria are met. A simple and popular formulation of this type is called an Upper Level Set (ULS) scan statistic \cite{patil2004upper, patil2006spatially, modarres2007hotspot}. ULS requires a tessellation of space (e.g., by geographic unit boundary; grid-partitioning), and starts by calculating a test statistic
for each cell. Then, a cell belongs to an upper level set of threshold $t$ if its test statistic value $\geq t$. All connected components in an upper level set form the set of candidate clusters, with potentially arbitrary shapes. A ULS tree is constructed by applying a series of thresholds and significant clusters are merged to generate final results. Similarly, a minimum-spanning-tree (MST) based approach enumerates connected sub-trees that are part of an MST of a tessellated input data, where the edge weights are inferred directly or indirectly by test statistic values \cite{costa2012constrained}. There also exists a FlexScan approach that enumerates all possible connected subsets in a tessellation in a brute-force manner \cite{tango2005flexibly, takahashi2008flexibly}. However, due to the exponential nature of the search space, it can only identify small clusters (e.g., <30 cells \cite{tango2005flexibly,Neill2015, duczmal2009extensions}). A trajectory based method was developed to relax the requirement of a tessellation \cite{demattei2007arbitrarily}, where data points are sequentially linked to form a trajectory using test statistic values; 
clusters are then enumerated by going through connected segments.

Meta-heuristics have also been explored to generate irregularly shaped candidates, and most still require a tessellated input space. A simulated annealing formulation \cite{duczmal2004simulated} was used to enumerate candidates by iteratively search over neighboring choices, each defined as another candidate that differs by one-cell from the current candidate. As a simulated annealing approach, the probability of moving into an inferior neighboring choice gradually decreases as the iteration number increases. Many genetic algorithm based formulations \cite{sahajpal2004applying, duczmal2007genetic} were also developed to generate arbitrarily shaped candidates, where off-springs are evaluated by a choice of test statistics. 
In addition, random-walk based method was developed to search free-form regions using a heuristic metric \cite{janeja2008random}.

To reduce over-complicated "tree shapes" of clusters that are generated as a result of the freedom of arbitrary shapes but are spatially not meaningful, regularization functions were developed to penalize such shapes based on geometric compactness defined by various criteria (e.g., ratio between a cluster's area to its convex hull \cite{duczmal2006evaluation, duczmal2007genetic}, sparsity \cite{bayesian1, shao2011generalized}). The regularization terms were also used in multi-objective formulations to generate pareto sets where clustering results are provided under different levels of shape-regularity constraints \cite{canccado2010penalized}.

While traditional clustering techniques have provided many efficient schemes to detect irregularly shaped clusters (e.g., DBSCAN \cite{dbscan}, OPTICS \cite{optics}, spectral clustering \cite{von2007tutorial}, Chameleon \cite{karypis1999chameleon}, HDBSCAN \cite{campello2015hierarchical, neto2017efficient}, etc.), these methods, as introduced earlier, do not incorporate statistical rigor required by domain applications of spatial hotspot detection. Recently, a statistically robust formulation of DBSCAN -- Significant DBSCAN -- was developed to explicitly control the rate of spurious clusters \cite{sigdbscan, sigdbplus}, though it currently does not model the distribution of control points described in Table \ref{tab:data}, Sec. \ref{sec:obs}.

\subsubsection{Constraints}\label{sec:constraint}

There are two types of constraints widely used for spatial hotspot detection. The first type concerns mostly about computational feasibility, i.e., to mathematically constrain the enumeration space to a finite and tractable set of candidates (e.g., non-exponential); whereas the second type of constraints is used to improve solution quality.

\textbf{Computation-driven constraints:} Given that even an enumeration space of predefined shapes may have infinite number of candidates, techniques rely on constrained parameterization of shapes to make the enumeration feasible. In the Euclidean space, circles are often constrained by two points -- one at the center and the other on the circumference, corresponding to a $O(N^2)$ search space \cite{satscan}. A center point may come from the set of event incidents (e.g., disease cases), underlying control data (e.g., population points), or an additional input set of candidate centers (e.g., centers of cells in a grid \cite{circle2}). This two-point definition has a natural explanation in epidemiology or similar domains, where the spread of an event (e.g., an infectious disease) propagates from a center source. Similarly, a ring-shaped cluster is parameterized by two concentric circles \cite{ring, ring1}. Since the inner circle may have a low density of points (e.g., serial criminals often do not commit crimes too close to their residence locations), its enumeration space uses three-point circles where all three points lie on the circumference. The outer circle is still defined by one-point on the circumference as the center location has been determined by the inner circle; this leads to a $O(N^4)$ search space. For rectangular shapes, candidates are constrained to rectangles whose boundaries on all sides touch at least one data point or grid cell, yielding the same $O(N^4)$ space \cite{rect1, rect2, walther2010optimal}. Many other similar constraints are used for ellipses, non-orthogonal ellipses, etc. Since flexibly shaped patterns can hardly be parameterized, there is often no further computation-driven constraints beyond the implicit constraints implied by different heuristic search strategies (e.g., local constraints).

In the network space, shortest-path is a common constraint for linear path enumeration \cite{linear3, linear4, de2011detection}. While this may be a reasonable approximation for people's travel behaviors, many real-world paths
such as bus routes do not always follow shortest-paths. Instead, a relaxed simple-path constraint -- any path without self-intersections -- quickly increases the potential space to $O(N!)$ \cite{linear5}. Another extension beyond shortest paths is to directly define the enumeration space using real-world trajectories and their connected sub-components \cite{li2020significant}. For network-distance based circles \cite{tang2017detecting}, rings \cite{ring2}, etc., the same shortest-path constraint is used to define the space that is reachable from a center (e.g., by a threshold $r$).

\textbf{Solution-quality-driven constraints:} This type of constraints are developed to further reduce non-meaningful outputs beyond statistical modeling, and is irrespective of computational considerations. One most important constraint is the spatial contiguity constraint, requiring that any two locations inside a cluster are mutually reachable by its inner space. This typically increases the computational cost \cite{minato2019fast}; in fact, a linear algorithm exists if this constraint is taken out \cite{neill2012fast}.
A relaxed version of this contiguity constraint is the proximity constraint \cite{speakman2013dynamic, speakman2015scalable, speakman2016penalized, neill2012fast}, which requires segments inside a candidate to be within a local neighborhood defined either by a spatial distance or $k$-nearest neighbors; the segments are allowed to be spatially disjoint. Similarly, density-contiguity within each hotspot can also be enforced via local constraints \cite{sigdbscan, sigdbplus}. In network space, contiguity constraints are commonly defined by sub-graph connectivity or its relaxed formulations \cite{cadena2018graph, aksoylar2017connected}. 
Another widely used constraint is a maximum population threshold, which eliminates clusters that cover more than 50\% of underlying population used in control data (e.g., \cite{satscan}); otherwise, a cluster is considered as a general phenomenon rather than a "hotspot". As discussed in Sec. \ref{sec:enumspace}, regularity constraints were also used in arbitrarily-shaped cluster detection to avoid over complicated tree-like patterns \cite{costa2012constrained, canccado2010penalized, duczmal2006evaluation, duczmal2007genetic}.

\begin{table}
	\centering
	\small
	\caption{Region Enumeration and Maximization.}
	\label{tab:enum}
	\begin{threeparttable}
	\begin{tabular}{p{2cm}p{2.6cm}p{7.4cm}}
		\hline
		\makecell[c]{\textbf{Classification}\\\textbf{criteria}} & \hfil \textbf{Categories} & \hfil \textbf{Examples}\\
		\hline
		\vspace{-11.9pt}
		\hfil\multirow{2}{*}{\vspace{2.9pt}Definition}
		& 
		\makecell[c]{Pre-defined shape}
		& 
		\vspace{-6pt}
		\begin{itemize}[noitemsep,nosep,topsep=0pt,,partopsep=0pt,leftmargin=*]
			\item Euclidean: Circles \cite{satscan, lord2020investigation, greene2016daily, circle2, matheny2020scalable}, ellipses \cite{ellip1}, rectangles \cite{rect1, rect2}, rings \cite{ring, ring1}
			\item Network: Linear paths \cite{linear2, linear3, linear4, de2011detection, linear5}, sub-networks with shape constraints \cite{ring2, tang2017detecting}\vspace{-10.9pt}
			
		\end{itemize}\\
		\cline{2-3}
		&
		\makecell[c]{Flexible shape} &
		\vspace{-6pt}
		\begin{itemize}[noitemsep,nosep,topsep=0pt,partopsep=0pt,leftmargin=*]
			\item Problem-specific heuristics: ULS \cite{patil2004upper, patil2006spatially, modarres2007hotspot}, MST \cite{costa2012constrained}, FlexScan \cite{tango2005flexibly, takahashi2008flexibly}, trajectory \cite{demattei2007arbitrarily}
			\item Meta-heuristics: Genetic algorithm \cite{sahajpal2004applying, duczmal2007genetic}, simulated annealing \cite{duczmal2004simulated}, random walk \cite{janeja2008random}
			\item General clustering based \cite{sigdbscan, sigdbplus}
			\vspace{-10.9pt}
		\end{itemize}\\
		\hline
		\vspace{-11.9pt}
		\hfil\multirow{2}{*}{\vspace{3pt}\makecell[c]{Constraints}}
		&
		\vspace{-18.7pt}\makecell[c]{Computation-driven} & 
		\vspace{-6pt}
		\begin{itemize}[noitemsep,nosep,topsep=0pt,partopsep=0pt,leftmargin=*]
			\item Euclidean: Finite enumeration space by data points (e.g., two-point circles, concentric four-point rings) \cite{satscan, ring, ring1, rect1, rect2, walther2010optimal}
			\item Network: Finite space plus shortest paths \cite{linear3, linear4, de2011detection, tang2017detecting}, simple paths \cite{linear5}, or real-world trajectories \cite{li2020significant}; relaxed sub-graph connectivity \cite{aksoylar2017connected}
			\vspace{-10.9pt}
		\end{itemize}
		\\
		\cline{2-3}
		&
		\makecell[c]{Quality-driven} & 
		\vspace{-6pt}
		\begin{itemize}[noitemsep,nosep,topsep=0pt,partopsep=0pt,leftmargin=*]
			\item Spatial contiguity/connectivity \cite{speakman2013dynamic, speakman2015scalable, speakman2016penalized, aksoylar2017connected, rozenshtein2014event, cadena2018graph}
			\item Density contiguity \cite{sigdbscan, sigdbplus}
			\item Shape regularity \cite{costa2012constrained, canccado2010penalized, duczmal2006evaluation, duczmal2007genetic}
			\item Maximum population \cite{satscan, ring1}
			\vspace{-10.9pt}
		\end{itemize}
		\\
		\hline
		\vspace{-16.9pt}
		\hfil\multirow{3}{*}{\vspace{3pt}\makecell[c]{Algorithms}}
		&
		\makecell[c]{Exact} & 
		\vspace{-6pt}
		\begin{itemize}[noitemsep,nosep,topsep=0pt,partopsep=0pt,leftmargin=*]
			\item Bound based \cite{rect1, rect2, neill2004detecting, ring, ring1, tang2015elliptical, linear4, linear5, sigdbscan}
			\item Incremental updates \cite{smi, bayesian1, neill}
			\item Linear-time subset scanning \cite{neill2012fast, neill2013fast, speakman2016penalized, chen2014non, cadena2018graph}
			\item Shortest-path-tree \cite{linear3, linear4}, FMGT \cite{linear5}
			\vspace{-10.9pt}
		\end{itemize}
		\\
		\cline{2-3}
		&
		\makecell[c]{Approximation}& 
		\vspace{-6pt}
		\begin{itemize}[noitemsep,nosep,topsep=0pt,partopsep=0pt,leftmargin=*]
			\item Approximation-ratio based \cite{agarwal2005hunting, agarwal2006spatial, matheny2020scalable, rozenshtein2014event,  zhou2019stochastic}
			\item Asymptotic-bound based \cite{rufibach2010block, walther2010optimal, rivera2013optimal}
			\vspace{-10.9pt}
		\end{itemize}
		\\
		\cline{2-3}
		&
		\makecell[c]{Heuristic}& 
		\vspace{-6pt}
		\begin{itemize}[noitemsep,nosep,topsep=0pt,partopsep=0pt,leftmargin=*]
			\item Data sampling \cite{matheny2016scalable}, space reduction \cite{ring1, smi}, meta-heuristics \cite{duczmal2004simulated, duczmal2007genetic, janeja2008random}\vspace{-10.9pt}
		\end{itemize}
		\\
		\hline
	\end{tabular}
	\end{threeparttable}
\end{table}

\subsubsection{Maximization algorithms}\label{sec:alg}
Statistically robust clustering techniques typically detect only the best cluster in each round, and remove it before searching for the next cluster to avoid shadowing effects between patterns (detailed in Sec. \ref{sec:revisit}). Thus, after an enumeration space is defined, maximization algorithms are used to identify the cluster candidate with the highest test statistic value. For example, test statistics $T(S)$ in Table \ref{tab:measure} can be used as objective functions, and $S$ represents a cluster candidate where cluster-specific inputs (e.g., counts of case $n$ and control $a$ for the Poisson-based test statistic) are calculated.
Many new computational techniques have been developed by the data mining community to reduce the cost of maximization. In the following, we summarize these acceleration algorithms in three broad categories: exact, approximation and heuristic algorithms.

\textbf{Exact algorithms:} The vast majority of maximization algorithms fall into this category, which means they can guarantee that the output cluster indeed achieves the maximum test statistic value within an explicitly defined search space (i.e., the cluster is the same as that of a brute-force algorithm). As a result, exact methods mainly focus on pre-defined shapes (Sec. \ref{sec:enumspace} and \ref{sec:constraint}). 

The initial algorithmic improvement was introduced by SaTScan \cite{software}.
The algorithm is only applicable to circular-shaped candidate enumeration, which pre-sorts all the points by distance to each circle center. The pre-sorting frees a candidate evaluation step from performing range queries to identify which case or control points are inside the candidate (inferred by the order after sorting). This reduces the overall enumeration and evaluation cost from $O(N^3)$ to $O(N^2\cdot\log N)$, assuming both cardinalities of center and circumference points are proportional to $N$. Since then, many significant improvements have been developed to further reduce the cost.

The first major type of strategies uses pruning, i.e., to eliminate subsets of enumeration spaces where no candidate may have the maximum test statistic value.
An overlap-kd-tree based approach was first developed for finding square-shaped clusters in a two-dimensional grid \cite{rect1}. Overlaps are added to original kd-tree partitions so that candidates that span across partitions are not missed. Then, a branch-and-bound pruning strategy is used to enumerate through the tree. Since only the optimal cluster is selected in each round, the existing maximum score is used as thresholds to prune branches with a smaller upper-bound score. Later works have extended the overlap-kd-tree and branch-and-bound strategies to cover rectangular shaped clusters \cite{rect2}, and for multi-dimensional data \cite{neill2004detecting}.
Similarly, filter-and-refine strategies were designed for various pre-defined shapes (e.g., circles \cite{circle2}, rings \cite{ring, ring1}, ellipses \cite{tang2015elliptical}, linear paths \cite{linear3, linear4, linear5}, iso-distance sub-networks \cite{tang2017detecting}) in non-tessellated input spaces. Leveraging that calculating the number of points in a sub-grid (e.g., $O(1)$ using an integral image \cite{sigdbscan}) is much faster than in a continuous space, these techniques typically use a discretized grid space to compute upper bounds of test statistic values of candidates in continuous spaces. A refining step is only carried out to check exact candidates if the upper bounds exceed a threshold (either user-specified or using the current maximum). 
These bounds are often designed specifically for each shape based on their characteristics. In addition, incremental update rules have also been developed to efficiently compute test statistic scores of neighboring candidates (e.g., $O(1)$ amortized cost \cite{smi, bayesian1, neill}); the update rules depend on specific test statistics.

Another widely adopted strategy is based on a linear-time subset scanning (LTSS) property first introduced in \cite{neill2012fast} for tessellated data. LTSS only applies to scenarios where components (e.g., grid cells) of a cluster are unconstrained by spatial contiguity, i.e., it can be any arbitrary subset of data elements. 
By relaxing the contiguity constraint,
LTSS states that, for many test statistic functions $F$, there exists an ordering function $G$, such that the subset with a cardinality of $k$ always reaches the maximum $F$ score if its members have the top-$k$ individual $G$ scores in the data \cite{neill2012fast}. 
Assuming a sorted order of individual components by $G$ has been given, LTSS allows finding the optimal subset in linear time out of an exponential enumeration space. It has been shown that the LTSS property can be satisfied by a variety of test statistic functions \cite{mcfowland2013fast, neill2013fast, speakman2016penalized}, including Kulldorff's scan statistic, expectation-based Poisson or Gaussian function, separable exponential families and many quasi-convex functions. The method has also been used in non-parametric scan statistics \cite{chen2014non}, GraphScan \cite{cadena2018graph}, etc. A few special-cases have been proven to make LTSS able to output contiguous subsets as clusters (e.g., non-existence of break-tires \cite{chen2014non}).
As mentioned above, one key limitation of LTSS is the lack of spatial contiguity in each cluster. To mitigate this, a proximity-constrained extension was developed \cite{speakman2013dynamic, speakman2015scalable}. Instead of applying LTSS on the entire data, it sequentially go through all data components (e.g., grid cells) and apply LTSS only on the $k$-nearest neighbors of each component, or neighbors within its $r$-radius neighborhood. Discontiguity is still allowed for subsets within each local neighborhood, but this avoids long-distance separation.

In the network space, shortest-path trees were used to efficiently enumerate candidate paths between all pairs of points \cite{linear3, linear4}. Since data points may add a large number of low-degree (two-degree) nodes to a network, dynamic segmentation was employed to construct the shortest-path tree with only original network nodes and then dynamically retrieve shortest paths between data points. For the all-simple-path enumeration space, the number of candidates can be factorial \cite{linear5}. Thus, more aggressive pruning methods were developed. Instead of directly using network nodes to build a simple-path tree, fragmentation-multi-graph-trees are used, where each tree node represents a sub-network and paths within each node can be dynamically computed and reused. 
Pruning and filter-and-refine strategies described above for Euclidean space based methods were also used in network-based hotpsot detection \cite{tang2017detecting}.
To handle high computational cost (e.g., NP-hardness \cite{cadena2018graph, aksoylar2017connected}, high-order combinatorials \cite{sharpnack2013changepoint, sharpnack2015detecting}) incurred by constraints on sub-graph connectivity, compactness or sparsity, methods often use relaxed reformulations such as semi-definite programming \cite{aksoylar2017connected} or spectral graph theory based relaxations \cite{sharpnack2013changepoint, sharpnack2015detecting}, where results (e.g., detection power) can be rigorously analyzed.

\textbf{Approximation and heuristic algorithms:} We discuss algorithms in these two categories together as it is hard to draw an exact boundary given the various definitions used by techniques. In computational theory, an approximation algorithm should guarantee that it always returns a solution $S_{apx}$ whose objective function value $F(S_{apx})$ (i.e., test statistic value for hotspot mapping) is within $\epsilon$ of the optimal solution $F(S_{opt})$. For hotspot mapping, one challenge in clearly defining "approximation" is that -- from an application perspective -- it is as important to know "where" and "how large" a hotspot is in addition to a $F$ value; this can be further complicated considering the large number of shapes used. In addition, this problem has broad interests by researchers from statistics and computer science, making the use of "approximation" more diversified.

A major difficulty in approximating the solution of spatial hotspots is due to the combinatorial nature of candidate enumeration in a finite and discrete space of parameters. To bridge this gap, an approximation scheme was provided in \cite{agarwal2005hunting} to search for the maximum-scoring region over a set of axis-parallel rectangles. This approach requires the scoring function $F$ to be a convex discrepancy function, and one key idea was to simplify the optimization using a set of linear functions to construct an approximation. Using Kulldorff's spatial scan statistic as an example of $F$, the approach was able to reduce the search cost from $O(N^4)$ to $P(\frac{1}{\epsilon}N^2\log^2 N)$, where $\epsilon$ is the gap to optimality. The linear-function based approximation was further improved in \cite{agarwal2006spatial} leveraging the fact that only the maximum-scoring region is needed for each round. It also provides a new approximation algorithm for grid-based input data with an $O(g^3\cdot poly(\log g, \frac{1}{\epsilon}))$ complexity, where $g$ is the number of columns and rows in a $g\times g$ grid. For rectangular shaped hotspot detection, an approximation-version of the exact brand-and-bound algorithm was also given in \cite{rect2}. This extension uses a relaxed instead of an exact upper-bound and thus allows more aggressive pruning. Probabilistic results were provided to describe how likely each individual relaxed-bound will remain correct, though no distance to global optimality was given.
A scalable sampling-based approach was also developed for trajectory data, which established approximation relationship among variable error bounds, cluster shape, sample size and execution time \cite{matheny2020scalable}; it also considered various ways to spatially sample and approximate the input trajectories.
In network space, approximation algorithms have also been developed by analyzing and using the submodularity of objective functions \cite{rozenshtein2014event}.
For sparsity constraints, efficient stochastic gradient descent algorithms have been developed, which have linear convergence rates for various objective functions \cite{zhou2019stochastic}.
Another type of approximation algorithms aims to maintain the optimality with respect to asymptotic detection power, guaranteeing that the approach will give the optimal solution with sufficiently large number of samples \cite{rufibach2010block, walther2010optimal, rivera2013optimal}. Related results were first achieved for one-dimension data, where regions are defined as intervals \cite{rufibach2010block}. This approximation only needs a $O(N\log N)$ search over the $O(N^2)$ enumeration space. The analysis has been later extended to two-dimensional space \cite{walther2010optimal}, where the best candidate in a set of $O(N^4)$ axis-parallel rectangles can be approximated by an economical set of rectangles, with an $O(N\log^2 N)$ cost. Additional assumptions typically used by these results include continuous surface of control data, independence between samples, data types and certain test statistic functions. 

Several other methods have a clearer heuristic nature. In ring-shaped hotspot detection, for example, a Delaunay-triangulation was used to generate guesses of ring centers instead of exhaustively going over centers of all three-point circles, reducing the center space from $O(N^3)$ to $O(N)$ \cite{ring1}. For multinomial-model-based hotspot detection (e.g., hotspots of high-diversity), a heuristic was developed to gradually narrow search volume as candidate sizes grow, based on the observation that candidate scores tend to converge (i.e., stabilize) as their sizes get larger \cite{smi}. A empirical sampling based approach was also developed to use only a subset of all data points to perform hotspot mapping \cite{matheny2016scalable}. Finally, the metaheuristics reviewed for flexible-shaped hotspot enumeration (e.g., \cite{duczmal2004simulated, duczmal2007genetic, janeja2008random}) can also be considered as heuristic-based optimization strategies. While heuristic approaches in general will not guarantee solution quality, many have demonstrated high power and performance through empirical results.

\begin{table}
	\centering
	\small
	\caption{Statistical hypothesis testing.}
	\label{tab:sig}
	\begin{threeparttable}
	\begin{tabular}{p{2cm}p{2.6cm}p{7.4cm}}
		\hline
		\hfil \textbf{Paradigms} & \hfil \textbf{Aspects} & \hfil \textbf{Examples}\\
		\hline
		\vspace{-11.9pt}
		\hfil\multirow{2}{*}{\vspace{2.9pt}Frequentist}
		& 
		\makecell[c]{Modeling}
		& 
		\vspace{-6.2pt}
		\begin{itemize}[noitemsep,nosep,topsep=0pt,,partopsep=0pt,leftmargin=*]
			\item Random locations \cite{satscan, rect2, ring, sigdbplus} or fixed location with random values \cite{smi, huang2009weighted, value1, value2}
			\item Scale-independent \cite{satscan, linear3, takahashi2008flexibly}, scale-dependent \cite{rufibach2010block, walther2010optimal, nnscan}, or unified \cite{unified} testing
			\item Bayesian as sub-models \cite{zhang2008bayesian, das2009detecting, mcfowland2013fast}
			\item Nonparametric empirical p-values \cite{chen2014non, neill2007nonparametric}
			\vspace{-10.9pt}
		\end{itemize}\\
		\cline{2-3}
		&
		\makecell[c]{Computation} &
		\vspace{-6pt}
		\begin{itemize}[noitemsep,nosep,topsep=0pt,partopsep=0pt,leftmargin=*]
			\item Monte-Carlo simulation accelerated by bounds \cite{rect2, ring, sigdbplus}, shared computation \cite{linear3, linear5}, reduction \cite{unified}, etc.
			\item Test statistic distribution approximation \cite{chen2002approximations, chen2009approximations, glaz2009scan}
			\vspace{-10.9pt}
		\end{itemize}\\
		\hline
		\vspace{-11.9pt}
		\hfil\multirow{2}{*}{\vspace{3pt}\makecell[c]{Bayesian}}
		&
		\makecell[c]{Modeling} & 
		\vspace{-6pt}
		\begin{itemize}[noitemsep,nosep,topsep=0pt,partopsep=0pt,leftmargin=*]
			\item Direct priors \cite{neill2007multivariate, bayesian1} or priors as parametric distributions \cite{makatchev2008learning, neill2018} 
			\item Direct posterior probabilities as results \cite{neill2005bayesian, neill2006bayesian, neill2007multivariate, bayesian1} or with added tests (e.g., empirical) \cite{bayesian2, shao2011generalized}
			\item Rank-based \cite{que2008multi, que2011rank}
			\vspace{-10.9pt}
		\end{itemize}
		\\
		\cline{2-3}
		&
		\makecell[c]{Computation} & 
		\vspace{-5.9pt}
		\begin{itemize}[noitemsep,nosep,topsep=0pt,partopsep=0pt,leftmargin=*]
			\item Fast subset sums \cite{neill2012fast, shao2011generalized, neill2011generalized}
			\item Conjugate priors \cite{neill2005bayesian}
			\vspace{-10.9pt}
		\end{itemize}
		\\
		\hline
	\end{tabular}
	\end{threeparttable}
\end{table}

\subsection{Hypothesis testing}\label{sec:test}
Hypothesis testing is the final step in each round of hotspot detection, which determines if the optimal cluster is a true hotspot or a spurious pattern.

\subsubsection{Region maximization: A Revisit}\label{sec:revisit}
There are two major reasons for selecting and testing only the optimal region in a single round \cite{satscan, rect2, unified, li2011spatial, zhang2010spatial, neill2018}. First and most importantly, returning only the optimal region allows robust control over the false positive rate given a significance level $\alpha$; otherwise, detecting multiple regions (e.g., candidates at different locations or with different sizes) at the same time may be susceptible to multi-testing, i.e., increasing the chance of having a type-I error. As suggested by many studies, this maximization can be viewed as an automatic adjustment of significance levels for different groups of candidates (e.g., candidates of the same size) \cite{rufibach2010block, walther2010optimal, rivera2013optimal}. The intuition can also be easily explained from a data perspective \cite{unified, smi}. Given $M$ datasets generated under a null hypothesis $H_0$, having a critical value -- a score threshold derived to determine a candidate's significance at the $\alpha$ level -- on the maximum score makes sure that a spurious cluster can only be falsely reported as a hotspot in smaller than $\alpha M$ datasets. In other words, to get a spurious detection in a dataset under $H_0$, its maximum score has to pass the threshold on the maximum score. The second reason is that this helps reduce the mutual shadow effect between true hotspots \cite{li2011spatial, zhang2010spatial}; e.g., eliminating a true hotspot from data will make the signal of a secondary true hotspot statistically stronger. 

In order to detect multiple clusters, a significant hotspot will be removed from data (both its case and control points), and the same detection and testing procedure will be repeated on the updated data. While this may increase the risk of reporting a spurious hotspot at the dataset-level, the effect is very slight as the execution of any following round is conditional on the detection of a significant cluster in the previous round (i.e., strict unidirectional dependence between rounds).
There have also been many studies trying to increase the detection power for multi-hotspot scenarios by adjusting test statistic functions \cite{li2011spatial, zhang2010spatial}.

\subsubsection{Two paradigms}
Both frequentist and Bayesian based paradigms have been explored to test the significance of a cluster.
The frequentist approach is used for both spatial and spatiotemporal hotspot detection. While the Bayesian variation was originally proposed and mainly used for the latter (e.g., emerging hotspot detection using time-series data over space), it can still be used if prior information is available. Table \ref{tab:sig} shows a summary of the two paradigms.

\begin{figure}
	\centering
	\includegraphics[scale=0.4]{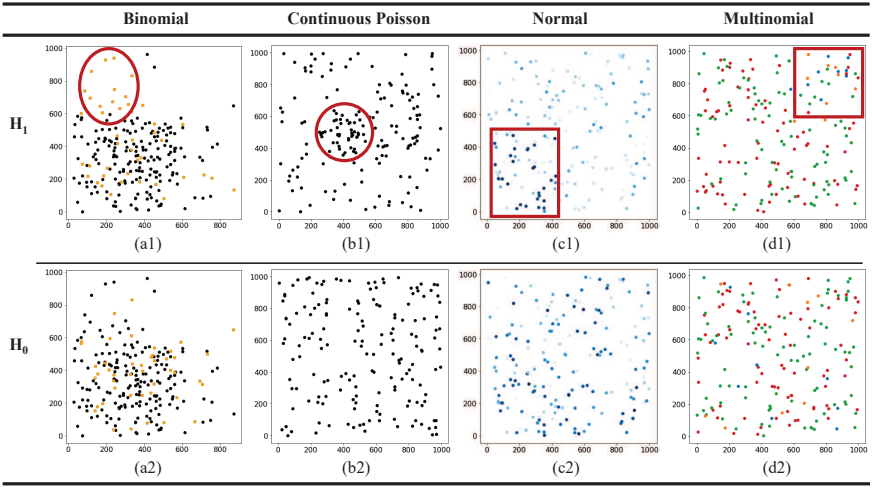}
	\caption{Example statistical models for spatial point processes (SPP) in Euclidean space. 
	The top-row shows observed data (i.e., input) with a true hotspot inserted using $H_1$; and the bottom row shows examples of simulation datasets generated by $H_0$ in a Monte Carlo trial. 
	Note that SPPs by default model the spatial distribution of points. While in this example point locations are allowed to be fixed (can be random), it can be interpreted as having the space confined to a set of spatial coordinates.
	}
	\centering
	\label{fig:mcs}
\end{figure}

\begin{figure}
	\centering
	\includegraphics[scale=0.346]{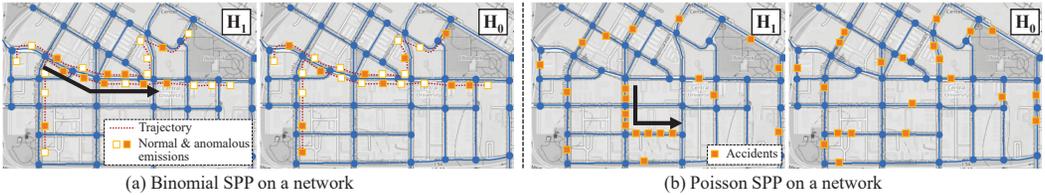}
	\caption{Examples of spatial point processes (SPP) in network space.
	}
	\centering
	\label{fig:network}
\end{figure}

\textbf{Frequentist Approach.} Frequentist based testing is the classic in this field and still dominates the use in related research and applications, partially because of the tradition and the popularity of the software SaTScan. 
The frequentist view considers the data as an aggregation of the past, so null hypotheses based on it typically assume an entire dataset is generated by a complete random process (e.g., homogeneous Poisson process).
The exact definitions of the hypotheses differ by statistical models discussed in Sec. \ref{sec:process}.

Fig. \ref{fig:mcs} shows several examples of random point distributions following different point processes in the Euclidean space.
Fig. \ref{fig:mcs} (a1) and (a2) show example datasets generated by Bernoulli-based spatial point processes (SPP) under $H_1$ and $H_0$, respectively; where the black and orange points are control and case points, respectively. Under the null hypothesis $H_0$,
each control point has an identical probability of being a case point (e.g., a COVID-19 case).  In contrast, $H_1$ in (a1) has a higher rate inside hotspot regions and lower outside. 
Fig. \ref{fig:mcs} (b1) and (b2) are example point distributions from the continuous Poisson point processes. Here $H_0$ assumes a homogeneous underlying probability density, which is also known as the complete spatial randomness. $H_1$ contains a region with higher probability density of generating points (i.e., a hotspot). 
In Fig. \ref{fig:mcs} (c2), $H_0$ of a normal-model-based point process assumes the values observed at all data points (e.g., PM-2.5 readings on air quality sensors) are generated by the same distribution. Instead, a hotspot in $H_1$ has a higher mean (of the normal model) as shown in Fig. \ref{fig:mcs} (c1). 
Finally, Fig. \ref{fig:mcs} (d1) and (d2) show example distributions from a multinomial-model-based point process, where $H_1$ in (d1) contains a region with significantly higher diversification of point types (e.g., biodiversity) and (d2) contains a random redistribution of the point types over space.
Fig. \ref{fig:network} further shows examples of Bernoulli-based and Poisson point processes in a network space with linear hotspots.

Based on the hypotheses, the test will produce a p-value for the optimal cluster $C^*_{obs}$ in the observed data $D_{obs}$:
$P_{H_0}(C^*_{obs},D_{obs}) = P\{F(C^*_{obs},D_{obs}) \leq F(C^*_{H_0}, D_{H_0})\}$, 
where $F$ is the test statistic function and $C^*_{H_0}$ is the optimal cluster in data $D_{H_0}$ following $H_0$.

Given the combined complexity of point processes, test statistics, enumeration spaces and maximization algorithms, the distribution of $F(C^*_{H_0}, D_{H_0})$ in general is unknown in closed-form, except in over-simplified special cases that are often not useful in practice \cite{rect2}. Thus, there have been two lines of strategies developed to estimate the p-value.

The most common strategy is to approximate the distribution via Monte Carlo simulation, i.e., generating a large number of randomized datasets (e.g., 999, 9999) following $H_0$ and running the same maximization algorithm on all datasets to simulate the distribution of $F(C^*_{H_0}, D_{H_0})$ (e.g., \cite{satscan, souza2019did, rufibach2010block, walther2010optimal}).
Fig. \ref{fig:mcs} (bottom row) shows randomization examples 
generated under $H_0$, where each can be considered as the simulated dataset in a Monte-Carlo trial.
Note that the spatial distribution of control points or base locations can either remain the same or be redistributed for Bernoulli, normal and multinomial point processes \cite{satscan, rect2, value1, smi}. In contrast, the control for continuous Poisson and many other processes is assumed to be a continuous surface, so the case point locations need to be redistributed in each trial.
This Monte Carlo approach is broadly applicable as long as $H_0$ is well defined, at the price of greatly increased computational cost.

In addition to the computational enhancements discussed in Sec. \ref{sec:alg}, which can be used as efficient sub-routines in Monte Carlo trials, a variety of dedicated algorithms have been developed to reduce the simulation cost (e.g., \cite{ring, linear4, rect2, neill2004detecting}). One convenient characteristic that has been leveraged by many approaches is that we do not need to know the exact maximum value $F(C^*_{H_0}, D_{H_0})$ for each random dataset $D_{H_0}$; instead, to calculate the p-value $P_{H_0}(C^*_{obs},D_{obs})$, all we need is whether $F(C^*_{obs},D_{obs})\leq F(C^*_{H_0}, D_{H_0})$ in the trials.
Using this characteristic, the maximum score from the observed data $F(C^*_{obs},D_{obs})$ is directly used as the bound in both branch-and-bound \cite{neill2004detecting} or filter-and-refine algorithms \cite{ring, linear3, linear4, linear5, li2020significant} to aggressively narrow search spaces. In addition, once a candidate is found to have a greater score than $F(C^*_{obs},D_{obs})$, the rest of the trial can be skipped. This also allows the use of non-exact but low-cost data structures during the simulation. As an example, a discrete-DBSCAN was used to replace the original DBSCAN for Monte Carlo trials to pre-estimate upper and lower bounds of $F(C^*_{H_0}, D_{H_0})$ \cite{sigdbscan, sigdbplus}; the original DBSCAN is only re-applied for a subset of trials where the bounds are insufficient to deduce the relationship.
Another widely adopted trick is early-termination \cite{software}. As its name suggests, this method terminates the whole Monte Carlo simulation once the results from completed trials are sufficient to conclude the significance. This is extremely efficient if a detection is insignificant. For example, for a significance level $\alpha=0.01$, the minimum number of trials needed to accept $H_0$ is 9 out of 999. It was also shown in \cite{sigdbscan} that early-termination is highly efficient for general hotspot detection formulations.

Shared computation has also been used for tasks that remain stationary across Monte Carlo trials. For example, many expensive operations for candidate enumeration and maximization (Sec. \ref{sec:alg})
only need to be pre-computed once,
such as the sorting of points \cite{value1, value2, value4, smi}, and shortest-path trees \cite{linear3, linear4} or fragmentation-multi-graph trees \cite{linear5} constructed for underlying road networks. 
The possibility of using only a small set of samples from the entire dataset to construct upper bounds of p-values has been explored. Specifically, a reduction algorithm was developed in \cite{unified}. It shows that the original significance-enforced clustering problem can be mapped to a series of reduced problems (e.g., from $N$ to $\lceil\rho N\rceil$ points) where the p-values of reduced clusters in these problems can be used to upper-bound the p-values of clusters in the original problem. Moreover, in scenarios where significance testing needs to be performed simultaneously for a collection of clusters, a dual-convergence algorithm was used to gradually narrow the gap between upper and lower bounds in pruning as the simulation progresses \cite{sigdbscan, sigdbplus}.

In addition to Monte Carlo simulation, another line of efforts has explored a variety of mathematical approximations of the distribution of $F(C^*_{H_0}, D_{H_0})$. Similar to many of the approximation algorithms summarized in Sec. \ref{sec:alg}, the approximation schemes here often aim to asymptotically approximate the distribution, i.e., narrowing the error bound as the number of samples gradually grows to infinity. Related work started on one-dimensional data \cite{naus1966some, glaz1991tight} or fixed-cluster-size (e.g., equal-size 1D or 2D windows \cite{chen2002approximations, chen1996two}) due to the complexity of analysis, and was later extended to two- and multi-dimension spaces for several types of test statistics (e.g., maximum normalized scan score) \cite{chen2009approximations, glaz2004multiple, glaz2006maximum}. Strategies have also been explored to select best-performing parameters in the approximation models \cite{chen2002approximations}, and many approximations have demonstrated good performance through comparisons with simulated results \cite{chen2002approximations, glaz2004multiple}. As discussed in Sec. \ref{sec:intro}, a detailed summary of these mathematical approximations has been provided in \cite{glaz2009scan}. Compared to the Monte-Carlo based techniques, accurate approximations of test statistic distributions in closed-forms may greatly reduce the computational overhead. On the other hand, Monte Carlo simulation tends to be more stable (e.g., fewer assumptions) and more flexible (e.g., can be generally applied to hotspot detection problems). Some efforts have also tried to combine Monte Carlo simulation with mathematical approximation as a middle ground between the two types of approaches \cite{software}. Finally, as discussed in Sec. \ref{sec:design}, p-values are commonly calculated in a scale-independent manner (i.e., the same critical value for all hotspot sizes), but scale-aware methods tend to be more robust \cite{rufibach2010block, walther2010optimal, unified}.

\textbf{Bayesian Approach:} Bayesian based testing schemes were developed to better model uncertainty during hotspot detection by incorporating informative prior information (e.g., expected number of observations in a spatial region, shape regularity, and size) when available.
As explained in \cite{neill2018}, prior information may also be added into frequentist approaches as constraints, weights or penalty functions in test statistics \cite{speakman2016penalized, rivera2013optimal, canccado2010penalized}, but the Bayesian approach offers more flexibility (e.g., priors as continuous distributions).
Moreover, a fundamental difference between the two frameworks is that the frequentist approach tests the significance of clusters by checking if $H_0$ can be rejected, whereas the Bayesian method (e.g., \cite{neill2005bayesian}) compares the posterior probabilities of the null and a set of different alternative hypotheses. In the multivariate Bayesian scan statistic \cite{neill2007multivariate}, for example, each alternative hypothesis $H_1$ is defined by a pair of a spatial footprint $S$ (i.e., a candidate region) and an event type $E$ (e.g., an increase in pharmacy visits may be caused by a disease outbreak, promotional sales of OTC medicines, etc.). A posterior probability is calculated for each hypothesis as $P(H_0|D) = \frac{P(D|H_0)\cdot P(H_0)}{P(D)}$ and $P(H_1(S,E)|D) = \frac{P(D|H_1(S,E))\cdot P(H_1(S,E))}{P(D)}$, 
where $D$ is the observed data, $H_0$ is the null hypothesis and $H_1$ is the alternative hypothesis for a region $S$ and event $E$ (there is typically a large number of ($S,E$) pairs).

Informative prior distributions and their parameters over space are the most critical components to calculate the large number of posteriors in Bayesian scan statistics \cite{neill2005bayesian, neill2006bayesian, neill2007multivariate, bayesian1}. 
Uninformative priors may reduce the Bayesian approach to certain versions of frequentist approaches such as the spatial scan statistic where candidates are ranked by their likelihood ratios \cite{neill2018} (many other types of test statistic functions exist as summarized in Sec. \ref{sec:measures}).

Original forms of Bayesian scan statistics methods \cite{neill2007multivariate, bayesian1} estimate the priors by directly learning their distributions (e.g., a prior conditional probability of an event occurring in each region) from past data within a relevant temporal window. Due to the large number of candidate regions (Sec. \ref{sec:enumspace}), this nonparametric learning requires a large amount of training data to work. To mitigate this issue, some of the prior distributions are further abstracted to parametric distributions that are governed by a small number of parameters \cite{makatchev2008learning}, which can be learned well with a smaller amount of training data. The downside is that the training is done through a computationally expensive expectation-maximization process \cite{neill2018}. In addition, the Bayesian approach normalizes the individual posteriors using the overall probability of data. This is again costly as $P(D) = P(D|H_0)P(H_0) + \sum_{S,E} P(D|H_1(S,E))P(H_1(S,E))$, which means all combinations of candidate regions and events need to be enumerated to calculate the summation. As a result, this method is constrained to simple regular shapes (e.g., circles, rectangles).

A remedy is the fast subset sums approach \cite{shao2011generalized, neill2011generalized}, which is an alternative to the above region-based Bayesian scan statistics. This method focuses more on individual locations and subsets of locations rather than contiguous regions. 
It was found in \cite{bayesian2} that the computationally-expensive terms in estimating the final posterior probability for each individual location $loc$, such as the sums of posteriors of all location-subsets that contain $loc$, can be exactly computed without enumerating through all the subsets. 
The result becomes a map of posterior probabilities on individual locations.
To improve the spatial proximity of locations with high posterior probabilities, a generalized fast subset sums method was developed \cite{shao2011generalized}, in which the proximity and regularity can be controlled by a new sparsity parameter.
Another computational enhancement used in Bayesian approaches is the construction of a conjugate prior \cite{neill2005bayesian}; otherwise expensive simulations such as Markov Chain Monte Carlo (MCMC) are needed to calculate the posterior.

The randomization test summarized in the frequentist approach is not used here for significance testing. Instead, the posterior probabilities (e.g., a regular-shaped cluster with highest posterior for $H_1$ or a map of posterior probabilities) are directly given to users for decision making. To enforce a false positive rate more explicitly, historical and synthetic data were used in \cite{bayesian2, shao2011generalized} to select a proper threshold on the posterior probabilities or its sums, where the number of false positives under different thresholds can be empirically evaluated. 

Many variations have also been developed. A Bayesian network model was developed to combine the spatial-based multivariate Bayesian scan statistic \cite{bayesian1} and the population-based PANDA \cite{cooper2007bayesian} to better capture details in individual data with spatial consideration \cite{jiang2010bayesian}. A temporal extension 
was later proposed \cite{bayesian4}, where the increase of density is modeled using a linear function. A rank-based spatial clustering algorithm uses the Bayesian scan statistic \cite{que2008multi, que2011rank} to compute posterior probabilities on individual locations, which were then used to search for clusters. A dedicated Bernoulli version of the Bayesian scan statistic was developed to improve performance on binary labeled points (e.g., a crime case or not, which is not aggregated) \cite{read2011bayesian}.

Integrated Bayesian-frequentist based approaches have been explored. A grid-based Bayesian variable window approach
first uses Bayesian to define the test statistic and then employs a frequentist based randomization test to perform significance testing \cite{zhang2008bayesian}. Similarly, methods such as anomalous group detection \cite{das2009detecting} and fast generalized subset scan \cite{mcfowland2013fast} use a Bayesian network to evaluate and select the best candidates, and randomization tests to compute critical values.

\subsection{Evaluation}\label{sec:eval}
One additional benefit of this statistically-robust clustering formulation is that ground truth hotspots are mathematically well-defined using point-process-based hypotheses (Sec. \ref{sec:test}). This allows rigorous synthetic data generation, where true hotpots with varying properties can be inserted into point processes to generate validation datasets (e.g., Fig. \ref{fig:mcs} and \ref{fig:network}). 
Such datasets have been widely used to evaluate results from different clustering approaches \cite{duczmal2006evaluation, disease, neill2009empirical, aamodt2006simulation, kulldorff2003power, huang2008evaluating, rivera2013optimal} under a variety of assumptions (e.g., data size, effect size, spatial scale). 
Typical metrics used in the literature include detection power (i.e., 1-$\beta$ where $\beta$ is the Type-II error) \cite{duczmal2006evaluation, neill2009empirical, kulldorff2003power}; rate of spurious results (i.e., Type-I error) or total number of spurious clusters detected \cite{neill2007multivariate, unified}; precision, recall and F1-scores \cite{nnscan, unified, smi}; and the number of days in advance in detecting an emerging hotspot in the spatio-temporal case \cite{neill2007multivariate, bayesian1, bayesian5, jiang2010bayesian, shao2011generalized}.
While counting the number of spurious detections in $H_0$-data is straightforward, measuring if a true hotspot is successfully detected can be tricky. Different strategies have been used to conduct this. For tessellated data (e.g., grid or polygon inputs), the success is often measured by the overlap ratio (i.e., intersection over union -- IoU) between the true pattern and detection \cite{jiang2010bayesian, shao2011generalized, speakman2013dynamic}. This can also be generalized to the IoU between point sets \cite{sigdbplus}. At the object level, IoU may be less robust. For example, the IoU between two circles decreases quickly even with small differences in center locations and radii \cite{unified}. Thus, separate criteria on center accuracy -- which may be important in applications (e.g., source of infection or pollution \cite{satscan}; locations of serial criminals \cite{ring}) -- and size similarity have been used as replacements \cite{unified, smi}. 
To make the validation process more holistic, incompatible assumptions have been used to generate synthetic data and test the sensitivity of techniques (e.g., regular vs. irregular shapes \cite{duczmal2006evaluation, costa2012constrained, tango2005flexibly}, normal vs. non-normal processes for continuous-value based clustering \cite{value1, value2, value4}).
In addition to quantitative evaluation, qualitative evaluation (e.g., real or simulated case studies, visual inspection) is also commonly employed \cite{patil2004upper, duczmal2006evaluation, ring1, linear3, linear4, sigdbscan, sigdbplus}.

In general, many empirical results (e.g., \cite{rufibach2010block, walther2010optimal, rect2, nnscan, unified}) have shown that spatial hotspots are easier to detect with larger data sizes, effect sizes and spatial footprints; fewer number of true hotspots; less uncertainty (e.g., due to aggregation or positional inaccuracy); regular shapes, etc. 
Related theoretical results have also been developed (e.g., detectability of a clustering signal) \cite{walther2010optimal, chan2013detection, proksch2018multiscale}.
Validation using real-world datasets (e.g., crime cases, traffic accidents, COVID-19 symptoms) often require deep involvement of domain experts. Many studies have used combinations of real-world base data (e.g., spatial domains such as city boundaries, county maps and road networks \cite{satscan, duczmal2006evaluation, kulldorff2003power, huang2008evaluating, linear3, linear4, linear5}; control data such as population or traffic volume \cite{rect2, bayesian1, li2020significant}) and synthetic event data (e.g., disease cases) in evaluation. Moreover, to further improve synthetic data's representativeness of real-world phenomenons, studies have gone beyond simple statistical assumptions to generate more realistic distributions (e.g., using imitation learning), such as synthetic population and contact networks \cite{zhao2015simnest, harland2012creating, bisset2009epifast}, human mobility maps \cite{bao2020covid}, trajectories \cite{pan2020xgail}, etc. These strategies for realistic spatial data generation can help better estimate the performance of techniques in real scenarios.
Finally, we list examples of open datasets across various domains that can be leveraged for evaluation and comparison in Table \ref{tab:eval}.

\begin{table}
	\centering
	\small
	\caption{Examples of open datasets for evaluation.}
	\label{tab:eval}
	\begin{threeparttable}
	\begin{tabular}{p{3cm}p{10cm}}
		\hline
		\makecell[c]{\textbf{Category}} & \makecell[c]{\textbf{Examples of open datasets}} \\
		\hline
		\makecell[c]{Simulated data with \\ground truth} 
		& 
		\vspace{-14pt}
		Northeastern US benchmark \cite{northeastern, kulldorff2003power, Song2003power, cadena2019near, diniz2020evaluating, cadena2017near, cadena2018graph};
		New York City disease outbreaks benchmark \cite{kulldorff2004benchmark, tango2005flexibly};
		arbitrarily-shaped (with customizable generation code) \cite{sigdbscan, sigdbplus}, etc.
		\vspace{4pt}
		\\
		\makecell[c]{Real data with known \\ground truth} 
		& 
		\vspace{-14pt}
		Battle of the water sensor networks \cite{cadena2019near, ostfeld2008battle, chen2016generalized, cadena2017near, cadena2018graph}; 
		New York Bronx Legionnaire’s disease outbreak \cite{ring1};
		Disease infection at airports (social network data) \cite{souza2019detecting},
		etc.
		\vspace{4pt}
		\\
		\makecell[c]{Real data without \\known ground truth$^*$}
		& 
		\vspace{-14pt}
		New York cancer incidence \cite{boscoe2016public, grekousis2021local, satscan_data},
		New York birth defects \cite{silva2021confidence, satscan_data},
        Sudden Infant Death Syndrome (SIDS) occurrences in North Carolina \cite{cressie1989spatial, minato2019fast};
        Metro Transit bus trajectories with engine measurements \cite{li2020significant};
        Los Angeles highway traffic \cite{cadena2019near, cadena2017near}; 
		Orlando pedestrians fatalities \cite{linear4},
		etc.
		\\
		\hline
	\end{tabular}
	\begin{tablenotes}[flushleft]
	\footnotesize
	\item $^*$The score of the best significant cluster and computational time are often used to compare methods, and domain 
	\item \hspace{2.1pt} knowledge is used to identify explanations and surprises.
	\end{tablenotes}
	\end{threeparttable}
\end{table}

\section{Future Opportunities}\label{sec:future}
There are many opportunities lying ahead to: (1) further enhance existing hotspot detection techniques, (2) incorporate statistical rigor into general clustering methods, (3) harness emerging problems and new forms of spatial data, and (4) explore synergistic integration with learning for mutual benefits.

\subsection{Improving solution quality, flexibility and efficiency}

As discussed in Sec. \ref{sec:measures}, the design of a test statistic is a critical component that directly determines the solution quality. Moreover, the design is non-trivial as localized test statistics (e.g., likelihood ratios \cite{satscan, rect1, ring}, density ratios \cite{linear1,linear3, linear4, linear5}, entropy measures \cite{smi}) can be easily biased when being used to compare and rank candidates in hotspot detection, especially across scales \cite{rufibach2010block, rivera2013optimal, walther2010optimal, chan2013detection}.
While approaches have been developed to correct the bias and restore the optimal detection power (e.g., \cite{rivera2013optimal, rufibach2010block, walther2010optimal, chan2013detection}),
most of them require sophisticated analysis and have only been explored in special cases of hotspot mapping (e.g., Gaussian white noises, one-dimensional intervals, rectangular regions). This limits the methods' adoption in general scenarios summarized in Sec. \ref{sec:main}, despite their importance. An interesting direction would be to explicitly explore the trade-off between flexibility and optimality, and develop more generally applicable frameworks that may not necessarily fully restore the optimal power.
In addition, recent methods that try to correct the bias by incorporating spatial-nondeterminism-awareness are less dependent on the specification of problems \cite{smi, nnscan, unified}, but they do introduce additional computational complexity and require more efficient algorithms and data structures.
Scalability-wise, existing algorithmic building blocks (Sec. \ref{sec:alg}) have not yet considered applications in the context of spatial big data \cite{parallel, yu2019spatial, evans2019enabling}. Enabling frameworks of large-scale applications need to be developed to model additional complexity (e.g., I/O, network communication, indexing) in distributed environments (e.g., Apache Sedona \cite{yu2019spatial}).

\subsection{More holistic modeling}
The trade-off between detection power and robustness against spurious patterns has not been sufficiently explored. While statistical rigor is certainly important to suppress false alarms -- especially in real-world applications where resource is limited, missing one true hotspot in many cases may turn out to be more costly in certain scenarios (e.g., the spread of COVID-19). Thus, to better inform decision making, an integration of the two indicators may be needed \cite{unified, bayesian1}. This integration can be non-trivial as the relative weights of detection power and false positive rate may be nonstationary during different stages of an event (e.g., disease spread) and can be affected by additional dynamic factors (e.g., cost of remedy, impact, policies, forecasts, geographical heterogeneity, local demographics). Consideration of these complex factors in hotspot recommendation may require measures beyond a single test statistic and simple criteria based on significance levels. These new modeling will also require new computational frameworks.

Moreover, the vast majority of techniques use the simplifying assumption that there is a single hotspot in the input data for $H_1$ \cite{li2011spatial, neill2018}. Although this does not necessarily limit the use of the approaches in detecting arbitrary number of clusters \cite{software, ring, unified, smi} (e.g., by performing rounds of detection until the best candidate is no longer significant, as discussed in Sec. \ref{sec:revisit}), it may cause the methods to work sub-optimally due to shadow-effects between multiple hotspots \cite{zhang2010spatial}. For example, existence of other hotspots increases the baseline score outside one true hotspot, making it harder to confirm its significance.
While several methods have been developed to mitigate this issue \cite{li2011spatial, zhang2010spatial}, they are again limited to certain special cases and have not been widely adopted (e.g., not explored for Bayesian modeling \cite{neill2018}). There are opportunities to more broadly consider this issue in hotspot detection and develop general frameworks to address it.

The uncertainty in input datasets also has not been well modeled in general except a few efforts (e.g., \cite{cadena2018graph, uncertain4, uncertain3}). Given that existing datasets in many problems (e.g., disease surveillance) are often provided at an aggregated level (e.g., by county or zip code) due to privacy concerns, such positional uncertainty of individual points can substantially affect solution quality \cite{uncertain1, uncertain5, uncertain6}. Explicit techniques (e.g., Dempster-Shafer based modeling) are needed in future research to improve the robustness of the techniques in face of uncertainty.

\subsection{Expanding to broad problems and emerging spatial data}

A major advantage of the wide variety of techniques summarized in this paper is their statistical robustness, which allows explicit control of the rate of spurious detections in the output. This modeling has also been used recently to improve the statistical robustness and automate parameter selection of general clustering techniques. Significant DBSCAN \cite{sigdbscan, sigdbplus}, for example, can effectively filter out a large number of spurious clusters generated by DBSCAN in data containing heavy-noise and clusters of varying densities. The incorporation of statistical modeling can also greatly reduce efforts in parameter selection. Similarly, a significant spatial data partitioning approach based on non-parallel split lines has been developed \cite{gaudart2005oblique}. However, in general, statistically robust formulations have not been explored for most general clustering and data partitioning approaches (e.g., HDBSCAN \cite{campello2015hierarchical, neto2017efficient}, spectral clustering \cite{von2007tutorial}, deep clustering \cite{min2018survey}, and many others \cite{xu2015comprehensive}). Many opportunities lie ahead to incorporate statistical rigor into these broad techniques.

In addition, there are many newly emerging spatial data types and problems where hotspot detection techniques may be valuable. As an example, location-enriched social network data (e.g., geo-tagged tweets) open up opportunities for monitoring disease spread and many other events (e.g., disasters) using novel formulations of hotspot detection \cite{souza2019detecting, souza2019did}. 
In the context of COVID-19, business points-of-interest data and their visits (e.g., SafeGraph \cite{safegraph}) can also be leveraged in hotspot detection to generate timely warning of high-risk regions and inform policy making.
The use of these new types of datasets in hotspot detection is also challenging due to well-known issues including data quality (e.g., positional accuracy and uncertainty \cite{uncertain5, uncertain6}), nonstationarity (e.g., algorithm dynamics, which are considered as major factors that cause erroneous predictions in Google Flu Trends in 2013 \cite{lazer2014parable}) and representativeness (e.g., selection bias \cite{kishore2020measuring}).

Many new types of data are also available in the network space. First, traditional GPS trajectories are currently underutilized in hotspot detection despite their potential importance (e.g., transportation safety, environment, smart cities, public health) \cite{datasource}. Second, attribute-enriched trajectories, including on-board-diagnostic data (i.e., vehicle trajectories with hundreds of real-time engine measurements \cite{li2018physics, li2020physics}) and connected vehicle data \cite{ali2015future}, also provide lots of new possibilities in novel hotspot detection formulations (e.g., routes with abnormally high emission or energy consumption \cite{li2020significant}). Also, many events in urban areas happen and propagate through road networks rather than the Euclidean space, so current research that are mostly focused on the Euclidean space may be expanded to the network space to take the full advantage of new high-resolution datasets. Advancements in these directions may require new statistical formulations (e.g., physics-aware point processes) and computational frameworks.

New opportunities also exist to go beyond existing hotspot formulations to address emerging societal problems. For example, the lack of diversity in many domains have raised major concerns in recent years on the resilience of our communities (e.g., low crop diversity in agriculture poses high risk to global food security \cite{pilling2020declining}). Research landscape in spatial hotspot detection can be broadened to help address these critical issues faced by our society \cite{smi}.

\subsection{Synergistic integration with learning}
One fundamental property of spatial data is spatial heterogeneity, i.e., data distributions are non-stationary over space, violating the common identical distribution assumption under machine learning methods. Recently, incorporation of statistically-robust formulations (e.g., multivariate scan statistics) in deep learning models has shown promising results in tackling spatial data heterogeneity and substantially improving model performances \cite{spatialnet, spatialstar}.

On the other hand, statistically robust clustering techniques can also leverage advances from related communities to improve its capacity.
For example, the growing maturity of deep learning \cite{lecun2015deep}, especially in computer vision and natural language processing, may provide a useful way to further enrich the variety of datasets that can be used for hotspot detection. While it can be risky to directly use deep learning to find hotspots due to concerns such as non-interpretability and overfitting issues \cite{castelvecchi2016can}, these models can help preprocess large volumes of visual and text information from remote sensing data (satellite or aerial imagery, LiDAR point clouds) \cite{jia2017incremental, xie2018timber}, social platforms (e.g., Twitter, Flickr), traffic cameras, connected vehicles, etc. to generate valuable new data sources for hotspot detection, which would otherwise be tedious, time-consuming and not scalable. Moreover, learning approaches (e.g., generative adversarial networks) may be leveraged to generate more realistic simulation data under various scenarios to improve evaluation \cite{zhao2015simnest, bao2020covid, pan2020xgail}.

From a computational perspective, data-driven approaches may also be used to reduce computation load and speed up the enumeration process \cite{luo2018surrogate} in scenarios where computing resource is limited on the application-side or real-time performance (e.g., interaction) is needed.
Machine learning methods, for example, may be used to heuristically prune low-potential enumeration spaces or recommend a list of high-potential regions for the search to prioritize \cite{zhai2019data, balcan2018learning}. Other settings such as reinforcement learning can also be helpful in this scenario.

To reduce the need of training data (e.g., in Bayesian hotspot detection techniques \cite{neill2007multivariate, bayesian1}), novel transfer learning methods \cite{tan2018survey} can be explored to reuse relevant parameters based on similarity between domain distributions across spatial locations, time periods, data or problems.

\section{Conclusions}\label{sec:conclusion}
In this article, we provided an overview of statistically robust clustering techniques for spatial hotspot mapping. Given the wide application of hotspot mapping techniques in societal applications and the potential importance (e.g., infectious disease surveillance for COVID-19 scenarios), the goal is to present a taxonomy of the vast literature in different key sub-fields (e.g., statistical modeling, maximization algorithms, significance testing) to trigger more thoughts from the broad computing communities. 
\footnote{Short summaries of the key steps are available in the online supplementary appendix for quick references.}
On the other hand, the statistically robust formulations may provide new paths for improving general clustering (e.g., removing spurious clusters) and learning techniques (e.g., heterogeneity); similarly, the validation strategies described (e.g., statistical definition based ground-truth generation) can be potentially used to expand existing evaluation methods for general clustering.
To serve these purposes, we selected the terms \textit{clustering} and \textit{hotspot} mainly due to their common use and adoption in the computing community. 
\footnote{Scan statistic has also been widely used in statistics communities and applications.}

We hope that the taxonomies and summaries of techniques provided in this article can serve as a stepping stone for generating new ideas and approaches (e.g., future directions discussed in Sec. \ref{sec:future}) in computing research, and help practitioners select appropriate paradigms based on their data and application needs.

\begin{acks}
This work is supported in part by the NSF under Grants No. 2105133, 2126474, 2126449, 1901099, 1737633, and 2040459, the USDOD under Grants HM0210-13-1-0005, ARPA-E under Grant No. DE-AR0000795, USDA under Grant No. 2017-51181-27222, NIH under Grant No. UL1 TR002494, KL2 TR002492 and TL1 TR002493, Google AI for Social Good, and the University of Maryland.
\end{acks}

\bibliographystyle{ACM-Reference-Format}
\bibliography{sample-base}

\newpage
\appendix
\section{Summaries on Key Building Blocks}

\subsection{Summary on statistical modeling}
There exists a variety of point processes (Table \ref{tab:process}) that can be used to model point distributions, and the choice often depends on specific application needs or assumptions. For data with time-series information or unknown statistical distribution, alternative formulations with Bayesian and non-parametric models can be leveraged.
The selection of test statistics requires more consideration. Specifically, scale sensitivity or bias (e.g., in density or likelihood ratio based formulations) should be explicitly considered when clusters of different sizes are being compared. Many correction methods (e.g., penalized or average likelihood ratios \cite{walther2010optimal, rufibach2010block, chan2013detection}) have been proposed for certain test statistics and formulations (e.g., rectangular shaped clusters, one-dimensional space). For more general formulations, flexible frameworks (e.g., dual-level Monte Carlo simulation \cite{smi}) can be employed with more computational cost.  

\begin{table}[h]
	\centering
	\footnotesize
	\caption{Examples of Statistical Models for Spatial and Spatiotemporal Point Processes.}
	\label{tab:process}
	\begin{tabular}{p{2.2cm}p{4.6cm}p{5.8cm}}
		\hline
		\makecell[c]{\textbf{Process types}} & \makecell[c]{\textbf{Suitable scenarios}} &  \makecell[c]{\textbf{Example use cases}}\\
		\hline
		\vspace{-5.3pt}
        \makecell[c]{Bernoulli or \\Poisson model}
		& 
		\vspace{-4pt}
		\begin{itemize}[noitemsep,nosep,topsep=0pt,partopsep=0pt,leftmargin=*]
			\item Sample values are counts (i.e., integers such as number of disease cases)
			\item Bernoulli also requires discrete geo-objects (e.g., points) \vspace{-10.9pt}
		\end{itemize}
		& 
		\vspace{-4pt}
		\begin{itemize}[noitemsep,nosep,topsep=0pt,partopsep=0pt,leftmargin=*]
			\item Identifying outbreaks of diseases such as legionnaire's disease \cite{ring1}, childhood leukemia \cite{leuk}, cancer (e.g., National Cancer Institute \cite{nih})
			\item Finding regions with abnormally high rates of crimes \cite{zeoli2014homicide}\vspace{-10.9pt}
		\end{itemize}\\
		\cline{2-3}
        \vspace{-5.3pt}
		\makecell[c]{Continuous\\ value model}
		& 
		\vspace{-4pt}
		\begin{itemize}[noitemsep,nosep,topsep=0pt,partopsep=0pt,leftmargin=*]
			\item Sample values are continuous (e.g., survival time, air pollution)\vspace{-10.9pt}
		\end{itemize}
		& 
		\vspace{-4pt}
		\begin{itemize}[noitemsep,nosep,topsep=0pt,partopsep=0pt,leftmargin=*]
			\item Identifying regions with significantly low birth-weights of infants,
			longer/shorter life expansions of residents, survival time after diagnosis of certain diseases \cite{value1, value2, value4, huang2007spatial, shen2014multivariate, huang2009weighted}
			\vspace{-10.9pt}
		\end{itemize}\\
		\cline{2-3}
		\vspace{-5.3pt}
		\makecell[c]{Multivariate\\ model}
		& 
		\vspace{-4pt}
		\begin{itemize}[noitemsep,nosep,topsep=0pt,partopsep=0pt,leftmargin=*]
			\item Samples are of different types that are related to a common event (e.g., different symptoms of a disease)\vspace{-10.9pt}
		\end{itemize}
		& 
		\vspace{-4pt}
		\begin{itemize}[noitemsep,nosep,topsep=0pt,partopsep=0pt,leftmargin=*]
			\item Identifying outbreaks of diseases where information from a single data source (e.g., one symptom) is an underrepresentation \cite{multi1, bayesian2, neill2013fast, neill2007multivariate}\vspace{-10.9pt}
		\end{itemize}\\
		\cline{2-3}
		\vspace{-5.3pt}
		\makecell[c]{Multinomial\\ model}
		& 
		\vspace{-4pt}
		\begin{itemize}[noitemsep,nosep,topsep=0pt,partopsep=0pt,leftmargin=*]
			\item Samples are of different types where each type is considered independent from others (e.g., different species or major types of diseases)
			\item There is no ordering among types\vspace{-10.9pt}
		\end{itemize}
		& 
		\vspace{-4pt}
		\begin{itemize}[noitemsep,nosep,topsep=0pt,partopsep=0pt,leftmargin=*]
			\item Finding sub-regions of a study area that have a different complexion of disease types \cite{multinomial}
			\item Finding vulnerable regions with low diversification of population (e.g., biodiversity, crop diversity, economic diversity) \cite{smi} \vspace{-10.9pt}
		\end{itemize}\\
		\cline{2-3}
		\makecell[c]{Expectation or \\Bayesian models}
		& 
		\vspace{-10pt}
		\begin{itemize}[noitemsep,nosep,topsep=0pt,partopsep=0pt,leftmargin=*]
			\item Each sample is a time-series
			\item Sample values are integer counts\vspace{-10.9pt}
		\end{itemize}
		& 
		\vspace{-10pt}
		\begin{itemize}[noitemsep,nosep,topsep=0pt,partopsep=0pt,leftmargin=*]
			\item Finding emerging hotspots of diseases or crimes \cite{neill2005detection, expectation, bayesian1, bayesian2} \vspace{-10.9pt}
		\end{itemize}\\
		\hline
	\end{tabular}
\end{table}

\subsection{Summary on region enumeration and maximization}
Both regular and irregular shape based formulations are common in literature. 
As regular-shaped clusters can be well-parameterized (e.g., two-point circles), their enumeration algorithms often use a top-down search strategy where candidates described by different parameters are directly enumerated. As a result, it is more robust if low-density gaps exist in true clusters, especially when the number of samples is limited. In contrast, irregular-shaped formulations often rely on a bottom-up approach using local search criteria; otherwise the enumeration space is often intractable. With that being said, fast linear algorithms have been proposed if spatial footprints of clusters are allowed to be dis-contiguous (e.g., with the LTSS property \cite{neill2012fast, speakman2016penalized}). 
In terms of maximization algorithms, 
approximation approaches tend to achieve lower worst-case complexity compared to exact algorithms. On the other hand, these results so far only exist for a limited set of hotspot detection problems, and often require substantially more efforts to migrate to other variations (e.g., arbitrarily shaped hotspots; evolving hotspots; hotspots based on trajectory data or other point processes; etc.). In contrast, the general computational paradigms (e.g., filter-and-refine, pruning) used for exact accelerations are easier to extend to cover different scenarios. 
Finally, heuristic algorithmic designs can be combined with both to further reduce the computational cost.

\subsection{Summary on hypothesis testing}
The comparison between frequentist and Bayesian approaches for hypothesis testing in statistically robust clustering inherits their philosophical differences in general (e.g., conclusions vs. probabilities for hypotheses, no-prior vs. prior, values vs. distributions for parameters, etc.).
In practice, the choice often depends on whether there exists a reasonable or agreeable prior distribution on parameters. If such a prior exists, then Bayesian may offer more flexibility such as probabilities on various hypotheses; otherwise, the inclusion of a prior may be considered subjective.
In terms of computation, algorithms for the frequentist paradigm often require randomization tests with Monte-Carlo simulation, except in special cases (e.g., certain choices of test statistics, cluster shapes, etc.) where approximations are available. Similarly, general Bayesian formulations may require expensive simulation (e.g., Markov-Chain Monte Carlo) unless suitable conjugate priors can be identified to directly estimate the posterior probability.

\end{document}